\pdfoutput=1
\documentclass{article} %
\usepackage{iclr2027_conference,times}

\usepackage[utf8]{inputenc} %
\usepackage[T1]{fontenc}    %
\usepackage[hidelinks]{hyperref} %
\usepackage{url}            %
\usepackage{booktabs}       %
\usepackage{amsmath}
\usepackage{amsfonts}       %
\usepackage{amssymb}
\usepackage{nicefrac}       %
\usepackage{microtype}      %
\usepackage{xcolor}         %
\usepackage{graphicx}
\usepackage{wrapfig}
\usepackage{makecell}
\usepackage{threeparttable}
\usepackage{multirow}
\usepackage{comment}
\usepackage{fontawesome}
\usepackage[noabbrev, capitalise]{cleveref}

\title{Every9D-22M: Large-Scale Real-World 9D\\Canonicalization of Everyday Objects}

\author{
Leonhard Sommer\thanks{Equal contribution.} \\
University of Freiburg
\And
Emil Akopyan$^*$ \\
University of Freiburg
\AND
Adam Kortylewski \\
CISPA Helmholtz Center for Information Security
}

\iclrfinalcopy %

\begin{document}

\maketitle

\begin{abstract}
Estimating the 9D pose of everyday objects from a single real-world image remains challenging. This is largely due to the lack of large-scale supervision. Most existing datasets either rely heavily on synthetic renderings or provide limited coverage of real-world objects. We address this gap with Every9D-22M, a dataset of 9D pose annotations for 22.8M real-world images from 114K object-centric videos spanning 630 everyday object categories, making it nearly two orders of magnitude larger than any prior real-world 9D pose dataset in at least one of category, object, or image count. 
To achieve this scale, we leverage object-centric videos by reconstructing object-level point clouds via multi-view geometry and aligning similar instances into a shared canonical coordinate frame. Canonical poses are manually annotated for only a small set of reference objects (fewer than $0.01\%$ of all images) and propagated to the remaining instances via cross-instance alignment. All propagated canonical poses are then verified from multiple viewpoints.
We further introduce cross-category orientation rules that induce category-level symmetries, enabling symmetry-aware evaluation. Beyond establishing dedicated training and evaluation splits as a benchmark for 9D pose foundation models, we show that training on Every9D-22M generalizes better to other datasets than training on existing datasets, and outperforms prior foundation models in rotation accuracy on all benchmarks except the viewpoint-biased ImageNet3D, where joint training with ImageNet3D surpasses them as well.
Data and code are available at \href{https://github.com/GenIntel/Every9D}{\faGithub/GenIntel/Every9D}.

\end{abstract}

\begin{figure}[!ht]
    \centering
    \includegraphics[width=0.90\linewidth]{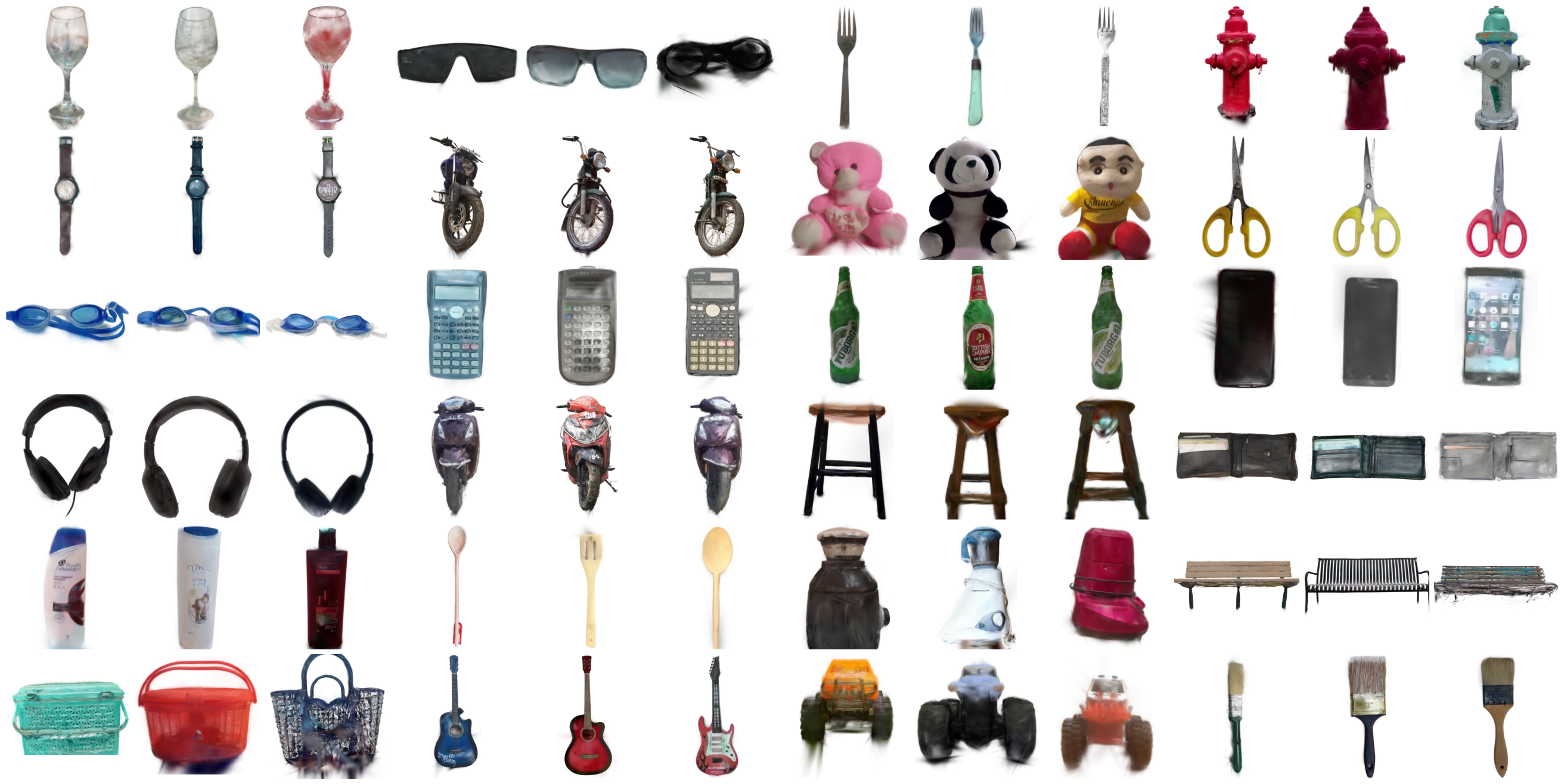}
    \caption{\textbf{Every9D-22M.} Example canonicalized objects from our dataset. Each object is reconstructed as a 3D Gaussian Splat from an object-centric video in uCO3D. Cross-instance alignment establishes shared canonical coordinate frames, which are subsequently propagated to all video frames, yielding 9D pose annotations for 22.8M real-world images across 630 object categories.}
    \label{fig:teaser}
\end{figure}
\section{Introduction}
Single-image 9D object pose estimation, comprising 3D rotation, 3D translation, and 3D object extent, is fundamental for robotics and AR/VR applications, as it enables spatial reasoning and physical interaction from a single RGB image. It requires models to infer consistent object-centric coordinate frames across diverse categories and instances. While recent methods have made progress in category-level 3D pose estimation~\citep{stark2010back, zhou2018starmap, xiao2021posecontrast, jesslen2024novum} and full 6D pose estimation~\citep{wang2019normalized, chen2021sgpa, lin2022single, liu2023net, zheng2023hs, zhang2024omni6dpose}, reliable 9D pose estimation for general everyday objects remains an open challenge under unconstrained real-world conditions.
Consistent object canonicalization also matters beyond robotics: text-to-3D and image-to-3D systems benefit from orientation-aligned training data, as canonical object frames improve controllability and geometric consistency~\citep{liu2024direct, zhang2024clay, lu2025orientation}, and correspondence learning and 3D morphable modeling commonly assume canonicalized objects to ensure stable alignment across instances~\citep{mariotti2024improving, dunkel2025yourself, choy20163d, groueix2018papier, park2019deepsdf, deng2021deformed, zheng2021deep, sun2022topology, sommer2025common3d}. Scalable real-world canonicalized supervision beyond synthetic data is therefore needed.

However, most existing large-scale 3D datasets rely on rendered assets~\citep{deitke2023objaverse, deitke2023objaversexl, zhang2025texverse, liu2023openshape, collins2022abo, khanna2024habitat, fu20213d}, which enable scale but are created under synthetic and controlled conditions. Models trained predominantly on such renderings, including the recent OrientAnything models for pose estimation~\citep{wang2024orient, wang2026orient}, struggle to generalize to real images affected by sensor noise, occlusions, motion blur, and appearance variability. To mitigate this sim-to-real gap, OrientAnything resorts to ImageNet3D~\citep{ma2024imagenet3d}, yet image-level 3D pose annotations are prohibitively expensive and fundamentally ambiguous from single views. Existing real-world pose datasets are thus forced to trade off category diversity against instance scale: those covering many categories provide only tens of thousands of annotated images, while those reaching millions of images cover only a handful of categories.

To overcome this limitation, we leverage object-centric videos as a scalable source of real-world supervision. For each video, we reconstruct an object-level point cloud via multi-view reconstruction, yielding a geometric representation robust to single-view ambiguities. We then cluster visually similar instances and align their point clouds through cross-instance geometric and appearance matching, establishing shared canonical coordinate frames that are propagated from a small set of manually annotated reference objects. Applying this framework to 114K object-centric videos of everyday objects from uCO3D~\citep{liu2025uncommon} yields Every9D-22M, with 9D pose annotations for 22.8M real-world images across 630 categories. Manual annotation covers fewer than $0.01\%$ of images, while every propagated pose is verified from multiple viewpoints, amounting to 199 hours of human effort in total: 67 hours of reference annotation and 132 hours of verification.

Our main contributions are the following:
\begin{enumerate}
    \item \textbf{Scalable real-world 9D canonicalization.} We introduce an iterative reference-based canonicalization framework that leverages object-centric videos for multi-view 3D reconstruction and cross-instance alignment. By propagating canonical frames from a small set of annotated reference objects, our approach enables 9D pose supervision at scale, with every propagated annotation subsequently verified from multiple viewpoints.
    \item \textbf{Every9D-22M: A large-scale real-world 9D pose dataset.} The resulting dataset covers 22.8M images across 630 categories, nearly two orders of magnitude more than any prior real-world 9D pose dataset in at least one of category, object, or image count. Fewer than 0.01\% of images require manual annotation.
    \item \textbf{Diverse large-scale benchmark for 9D pose foundation models.} We establish standardized symmetry-aware evaluation protocols and baseline models. Training on Every9D-22M generalizes better to other datasets than training on existing datasets and outperforms prior foundation models in rotation accuracy on all benchmarks except the viewpoint-biased ImageNet3D, where joint training with ImageNet3D surpasses them as well.
\end{enumerate}

\section{Related Work}
We review prior work on 9D object pose annotation from real-world imagery, organized by the level of annotation effort, and compare real-world 9D pose datasets quantitatively in \cref{tab:datasets}.

\begin{table}[b]
\centering
\caption{Comparison of real-world model-free pose datasets. $^\star$Note that depth estimates in our dataset are obtained using a monocular depth prediction method~\citep{yang2024depth} and are subsequently aligned with SfM. For Every9D-22M, we count only categories given five test instances.}
\footnotesize
\setlength{\tabcolsep}{2pt}
\renewcommand{\arraystretch}{1.15}
\label{tab:datasets}
\resizebox{\linewidth}{!}{%
\begin{tabular}{l | c  l r  r r r r r }
\toprule
Dataset & In-/Outdoor & Modality & Eval. & \#Categories & \#Objects & \#Videos & \#Images & \#Manual  \\
\midrule
PASCAL3D+~\citep{xiang2014beyond}            & \checkmark & RGB   & 3D & 12  & 12K   & /    & 12K    & 12K      \\
ObjectNet3D~\citep{xiang2016objectnet3d}     & \checkmark & RGB   & 3D & 100 & 57K   & /    & 57K    & 57K   \\
ImageNet3D~\citep{ma2024imagenet3d}          & \checkmark & RGB   & 3D & 200 & 86K   & /    & 86K    & 86K  \\
Objectron~\citep{ahmadyan2021objectron}      & \checkmark & RGB   & 9D & 9   & 17K   & 15K   & 4M    & 17K   \\
Wild6D~\citep{fu2022category}                & \checkmark & RGBD  & 9D & 5   & 162   & 486   & 10K   & 162   \\
HANDAL~\citep{guo2023handal}                 & \checkmark & RGBD  & 9D & 17  & 212   & 2K    & 308K  & 2K   \\
NOCS-REAL275~\citep{wang2019normalized}      & $\times$ & RGBD  & 9D & 6   & 42    & 18    & 8K    & ~90  \\ %
PhoCaL~\citep{wang2022phocal}                & $\times$ & RGBD  & 9D & 8   & 60    & 24    & 3.9K  & 60   \\
HouseCat6D~\citep{jung2024housecat6d}        & $\times$ & RGBD  & 9D & 10  & 194   & 41    & 23.5K & ~4.8K   \\ %
Omni6DPose-ROPE~\citep{zhang2024omni6dpose}  & $\times$ & RGBD  & 9D  & 149 & 581   & 363   & 332K  & ~1.8K   \\
Every9D-22M (Ours)        & \checkmark & RGBD$^\star$ & 9D & \textbf{630} & \textbf{114K}  & \textbf{114K}  & \textbf{22.8M} & 1K   \\
\bottomrule
\end{tabular}%
}
\end{table}

\subsection{Image-Level Pose Annotations}

Obtaining pose annotations at the image level is the most labor-intensive paradigm. Prior works~\citep{xiang2014beyond, xiang2016objectnet3d, ma2024imagenet3d} select an exemplary CAD mesh for a given category and manually align it to each image by adjusting its pose parameters. This is inherently sensitive to discrepancies between the template mesh and the true object: the projected alignment can appear accurate in the 2D image plane while remaining geometrically inconsistent in 3D. Given scale ambiguity and the absence of real-world camera intrinsics, these approaches evaluate only 3D object orientations. ImageNet3D~\citep{ma2024imagenet3d} is the largest such benchmark with approximately 86K images, whereas we scale to 22.8M annotated images with only around 1K manual pose annotations.

WildDet3D~\citep{huang2026wilddet3d} instead annotates 3D bounding boxes at the image level, with orientation defined primarily by the gravity direction, whereas we leverage semantically grounded cross-category intrinsic and interactive orientation rules, detailed in the supplementary material.

\subsection{Video-Level Pose Annotations}

Video-level annotation reduces ambiguity by exploiting multi-view consistency across frames. Where each video contains a single object instance, human effort scales with the number of videos, as in Objectron~\citep{ahmadyan2021objectron}, which annotates poses manually per video, and in Wild6D~\citep{fu2022category} and HANDAL~\citep{guo2023handal}, which propagate keyframe and tracker initializations and refine them manually. With multiple objects per scene and instances reappearing across videos, the effort scales with both quantities: NOCS-REAL275~\citep{wang2019normalized} relies entirely on manual labeling, while PhoCaL~\citep{wang2022phocal} and HouseCat6D~\citep{jung2024housecat6d} match previously scanned objects to scene observations using robot-arm keypoint measurements.

Labeling each instance within a sequence nevertheless remains labor-intensive, limiting scale. Objectron~\citep{ahmadyan2021objectron} is the largest dataset by number of annotated objects, comprising approximately 17K instances, but it covers only 9 categories, nearly two orders of magnitude fewer than our 630. Omni6DPose-ROPE~\citep{zhang2024omni6dpose}, the second largest by object instances, contains only 581 objects, again nearly two orders of magnitude fewer than our 114K.

uCO3D~\citep{liu2025uncommon} normalizes object-centric reconstructions at the scene level by enforcing a gravity-aligned vertical axis, normalizing scale and translation, and resolving horizontal orientation using PCA. This removes the arbitrary coordinate frames of structure-from-motion pipelines, but does not guarantee semantic object orientation: the vertical scene axis need not match the semantic top--bottom direction of the object, and PCA-based yaw alignment is not semantic either. In contrast, we canonicalize objects semantically, according to meaningful orientation rules shared across instances.
\subsection{Reference-Level Pose Annotations}

Methods exploiting automatic cross-instance alignment remain limited~\citep{sommer2024unsupervised, chi2026cpo, jin2025one}. The first method aligning instances to a single reference~\citep{sommer2024unsupervised} is limited to 1K videos across 23 categories, does not filter incorrect poses after alignment, and requires category information to select a reference. C3PO~\citep{chi2026cpo} canonicalizes 200 videos across 20 categories using category supervision, extending single-reference canonicalization to partially observed objects through pairwise all-to-all alignment, whose quadratic complexity limits scalability. More recently, \citet{jin2025one} canonicalize 32K filtered Objaverse assets using one annotated reference per category, again assuming category information; their support-plane strategy is questionable for the noisy, partial point clouds obtained from structure-from-motion with imperfect masks, and has not been shown to generalize to real-world videos. Our method requires no category information, allowing cross-instance alignment across categories and reducing human effort.

\subsection{Self-Supervised Pose Annotations}
MagicPony~\citep{wu2023magicpony} recovers 6D pose and shape self-supervised from real-world data, but targets articulated animals, reports no quantitative pose accuracy, and has not been shown to generalize to rigid everyday objects. Purely geometric approaches~\citep{li2021leveraging, sajnani2022condor} have so far demonstrated strong performance primarily on a limited set of categories, such as airplanes, cars, and bottles, and typically assume complete point cloud observations, with accuracy degrading substantially for partial ones; \citet{jin2025one} further shows that Condor~\citep{sajnani2022condor} performs particularly poorly in low-instance regimes. Given the long-tailed distribution of object instances in real-world data and the noise introduced by structure-from-motion reconstruction and imperfect object masks, these limitations become critical in our setting.

\section{Method}
\label{sec:method}

Our objective is large-scale, high-quality 9D pose annotation for real-world object-centric videos. To scale annotation while keeping orientation consistent across instances, we adopt a reference-based canonicalization strategy: a small set of manually annotated reference objects defines canonical coordinate frames that are propagated automatically to many instances and their video frames through cross-instance and multi-view alignment.

\begin{figure}[!b]
    \centering
    \includegraphics[width=\linewidth]{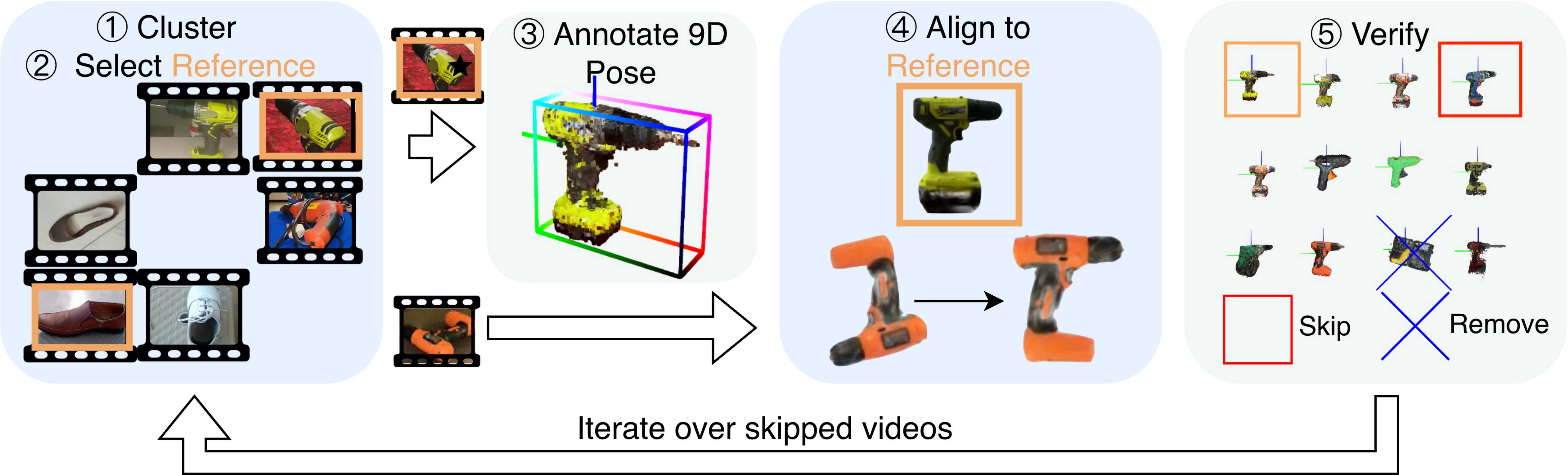}
    \caption{\textbf{Canonicalization Framework.} Our framework to canonicalize object-centric videos consists of 5 steps. 1) We cluster all object-centric videos. 2) We select a reference per cluster. 3) We annotate a 9D pose for the reference. 4) We align each object to its reference based upon geometry and appearance. 5) We verify all objects, removing objects with bad reconstruction and skipping objects with wrong pose annotation. Next, we repeat all steps over all skipped videos. Also note that manual labour is only required for step 3) 9D Pose annotation of the reference and 5) verification. }
    \label{fig:method_overview}
\end{figure}

Our framework comprises five stages, illustrated in \cref{fig:method_overview}: clustering videos (\cref{sec:method_clustering}), selecting a reference per cluster (\cref{sec:method_reference_selection}), annotating its 9D pose following our orientation categories (\cref{sec:method_pose_annotation}), aligning the cluster's remaining videos to that reference (\cref{sec:method_alignment}), and verifying the propagated annotations (\cref{sec:method_verification}); we then iterate over the videos left over (\cref{sec:method_iterative}). We additionally train a 9D pose estimation network as a baseline (\cref{sec:method_pose_estimation}).

\subsection{Clustering Videos}
\label{sec:method_clustering}

\begin{wraptable}{r}{0.51\textwidth}
\centering
\footnotesize
\setlength{\tabcolsep}{2pt}
\begin{threeparttable}
\caption{\textbf{Canonicalization design validation} on a small annotated subset of CO3D. We validate (a) the choice of a medoid reference over a random reference and (b) the benefit of multiple alignment iterations. Row~1 is UOP3D~\citep{sommer2024unsupervised}.}
\label{tab:ablation}
\begin{tabular}{c l c c c c c}
\midrule
\# & Method & Clustering & Ref. & Iter. & \makecell{GT /\\iter.} & \makecell{Acc@30$^\circ$\\$\uparrow$} \\
\midrule
1 & UOP3D                               & Category     & Random & --- & 10\% & 82.2\% \\
2 & Ours                                & DINOv3  & Random & --- & 10\% & 76.1\% \\
3 & Ours                                & DINOv3  & Medoid & --- & 10\% & 83.0\% \\
4 & Ours                                & DINOv3  & Medoid & 2 & 5\%  & 86.5\% \\
\bottomrule
\end{tabular}
\end{threeparttable}
\end{wraptable}

The inputs to this stage are object-centric videos recorded by non-experts, typically capturing the object from diverse viewpoints, often close to a full 360° around it. They vary substantially in distance, elevation, illumination, background clutter, and motion blur.

To assign videos to clusters sharing a common reference instance, we compute a single embedding per video. Specifically, we extract DINOv3~\citep{simeoni2025dinov3} features from 32 uniformly sampled frames and average them after normalization, which stays within 1.5 points of using 64 frames on the CO3D benchmark at half the cost. We then apply cosine $k$-means ($k=500$) to the normalized embeddings.

We validate clustering on a small CO3D subset~\citep{reizenstein2021common} with annotated 3D poses~\citep{guo2023handal}, see~\cref{tab:ablation}. Category-agnostic clustering nearly matches the accuracy of ground-truth category grouping, indicating near category-level cluster quality. Unlike category-based grouping, it lets us tune the number of clusters to control how many references need manual annotation, is robust to incorrect category labels, and exploits cross-category visual similarity. Accordingly, two alignment iterations on half the ground-truth annotations improve performance, even on this subset of 20 categories with 10 similar instances each.

\subsection{Reference Selection}
\label{sec:method_reference_selection}
Instead of a random reference~\citep{sommer2024unsupervised}, we choose the cluster medoid, i.e., the video whose image-feature embedding has the smallest average distance to all others in the cluster. This yields a more representative reference and significantly improves accuracy, see~\cref{tab:ablation}.

\subsection{9D Pose Annotation}
\label{sec:method_pose_annotation}

We first obtain camera poses and a sparse 3D point cloud with VGGSfM~\citep{wang2024vggsfm}, and extract it by multi-view masking~\citep{sommer2024unsupervised} using text-conditioned LangSAM~\citep{kirillov2023segment,medeiros2023langsegmentanything} masks stabilized by XMem~\citep{cheng2022xmem}.

Fitting a centered oriented 3D bounding box to the masked point cloud turns each orientation annotation into a 9D pose annotation. For precision, annotators use an interactive tool, placing axes coarsely and refining them in $1^\circ$ increments; all labels were cross-verified by a second annotator.

To define orientation consistently across instances, we establish cross-category rules (\cref{sec:sup_orientations}), grouping object categories into orientation categories (\cref{sec:sup_orientation_mapping}), based on three principles: (a) \emph{intrinsic orientation}, such as a mounting direction or the primary direction of motion; (b) \emph{human–object interaction}, whether an object faces toward or away from the user and how it is grasped; and (c) \emph{object–object interaction}, reflecting functional usage such as painting or squeezing direction. For example, the canonical front of a chair is opposite the backrest, while for scissors the forward axis aligns with the cutting direction. Orientation categories that leave axes underdetermined additionally induce symmetries, which we derive automatically: a bottle defines only \emph{contain upward}, giving a continuous symmetry about that axis, whereas a tray defines \emph{along grab two hands left right}, an axis rather than a direction, giving a discrete symmetry.

\subsection{Alignment to Reference}
\label{sec:method_alignment}

We estimate the rigid transformation aligning a reconstructed instance with its cluster's canonical reference, jointly leveraging geometric structure and visual appearance: each video is represented as a 3D surface endowed with DINO features~\citep{simeoni2025dinov3} projected onto a reconstructed mesh, following~\citep{sommer2024unsupervised}.

From the masked sparse reconstruction (\cref{sec:method_pose_annotation}), we estimate a coarse object surface via alpha shapes~\citep{edelsbrunner1994three} and project multi-view DINO features onto it, giving a surface feature representation $\mathcal{S} = \{\mathcal{V}, \mathcal{F}\}$, with vertices $\mathcal{V} = \{\mathbf{v}_i \in \mathbb{R}^3\}_{i=1}^{|\mathcal{V}|}$ and multi-view features 
$
    \mathcal{F} =
    \left\{
        \left\{
            \mathbf{f}_i^{\,k} \in \mathbb{R}^D
        \right\}_{k=1}^{|\mathcal{F}_i|}
    \right\}_{i=1}^{|\mathcal{V}|}.
$ 
Here, $\mathbf{f}_i^{\,k}$ is the $k$-th feature observation of vertex $i$.

We estimate the relative alignment $T$ between two surfaces $\mathcal{S}_i$ and $\mathcal{S}_j$, composed of rotation, translation and scale, using RANSAC initialization followed by gradient-based refinement, as in~\citep{sommer2024unsupervised}. The alignment minimizes a weighted combination of geometric and appearance distances:
$
    \mathrm{dist}(\mathcal{S}_i, \mathcal{S}_j)
    =
    (1 - \alpha)\,\mathrm{dist}_{\mathrm{geo}}(\mathcal{S}_i, \mathcal{S}_j)
    +
    \alpha\,\mathrm{dist}_{\mathrm{app}}(\mathcal{S}_i, \mathcal{S}_j),
$
where $\alpha \in [0,1]$ controls the trade-off between geometry and appearance. In practice, we set $\alpha=0.2$.

The geometric term $\mathrm{dist}_{\mathrm{geo}}$ is a weighted Chamfer distance between vertices; the appearance term $\mathrm{dist}_{\mathrm{app}}$ is evaluated in 3D but with nearest neighbors in feature rather than spatial proximity.

Specifically, for each vertex $i$, we identify the target vertex whose associated feature set yields the smallest aggregated nearest-neighbor feature distance:
$
        \mathbf{v}_{\mathrm{nn}}^{\mathrm{feat}}(i)
    =
    \arg\min_{j \in \{1,\dots,|\mathcal{V}_j|\}}
    \sum_{l}
    \min_{k}
    \left\|
        \mathbf{f}_j^{\,k}
        -
        \mathbf{f}_i^{\,l}
    \right\|_2,
$
for $i \in \{1,\dots,|\mathcal{V}_i|\}$. The bi-directional appearance distance is computed between the corresponding 3D vertex locations, each correspondence weighted by its feature-space cycle consistency~\citep{sommer2024unsupervised}. Finally, we fit an oriented 3D bounding box to the aligned masked point cloud and propagate the resulting 9D pose to all frames of the video.

\subsection{9D Pose Verification}
\label{sec:method_verification}
For each sequence, we visualize the canonicalized reconstruction from three canonical views (front/top/right) and check whether the pose matches the object's semantic axes and the reconstruction is of sufficient quality.

We categorize failures into two groups. \textbf{Filtered} ones are permanently removed because the geometry is unreliable, e.g., poor reconstruction (moving objects, noise, missing parts) or masks containing multiple instances. \textbf{Skipped} sequences are kept for later processing, as the failure stems from the reference rather than the geometry: overly diverse clusters, fine-grained state changes (e.g., open vs.\ closed scissors), or semantically different objects grouped together (e.g., a spoon aligned to a pan). As summarized in \cref{tab:verification_statistics}, 74.9\% of sequences are accepted across both iterations, while 18.1\% are filtered and 7.0\% remain skipped.

\subsection{Iterative Process}
\label{sec:method_iterative}
We apply the pipeline for two iterations: after each pass, we remove filtered sequences and re-process the skipped subset by re-clustering, selecting new references, annotating poses, and repeating alignment and verification. Overall, we annotate 1000 reference objects across both iterations.
\subsection{Monocular 9D Pose Estimation}
\label{sec:method_pose_estimation}
To demonstrate the usefulness of Every9D-22M for training monocular pose estimators, we train a simple baseline model for 9D pose prediction. Following prior work~\citep{wang2019densefusion, zhang2024omni6dpose, zhang2025beyond}, we fuse geometry and appearance by re-projecting each pixel into 3D using its depth and attaching its patch feature.
As point cloud backbone we employ LitePT~\citep{yue2025litept} rather than PointNet++~\citep{qi2017pointnet} or DGCNN~\citep{phan2018dgcnn,yu2021pointr,zhang2025beyond}, combining sparse 3D convolutions~\citep{graham20183d} with point transformer layers~\citep{wu2024point}, and encode images with a pretrained DINO backbone~\citep{oquab2023dinov2}. The features are concatenated and passed through a two-layer MLP predicting translation, size, and 6D rotation~\citep{zhou2019continuity}, trained with a geodesic rotation loss and $\ell_1$ losses on translation and extent.

\begin{figure}[t]
  \centering
  \includegraphics[width=\textwidth]{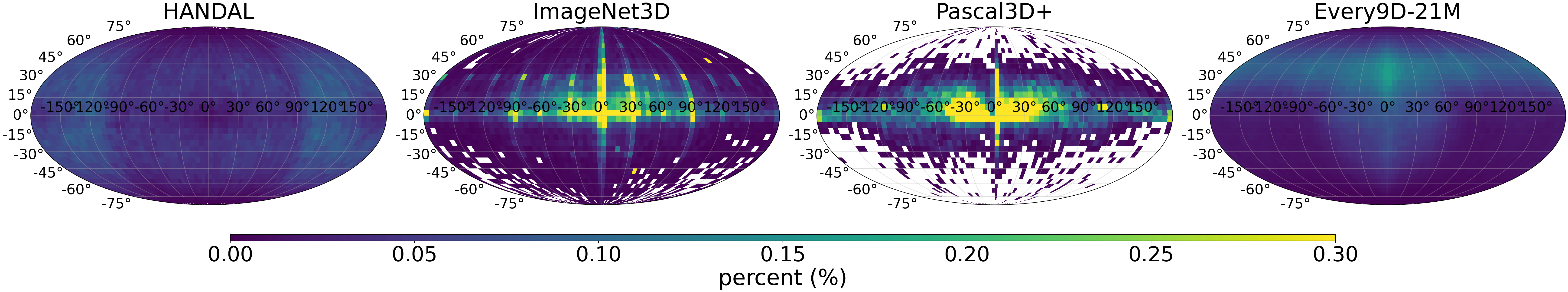}
  \caption{\textbf{ Mollweide projection of camera-direction histograms in the canonical object frame} Every9D-22M is the only dataset providing broad, near-uniform coverage of the viewing sphere; ImageNet3D and Pascal3D+ concentrate sharply along the equator (canonical photograph angles), while HANDAL covers the hemisphere uniformly but contains very few categories.}
  \label{fig:dataset_distr_viewpoint}
\end{figure}

\section{Dataset}
\label{sec:dataset}

We present a comprehensive overview of our dataset, which comprises 22.8M images, 114K annotated objects, and 630 categories with five test instances each across many super-categories. The images are obtained by sampling 200 frames per video. In practice, given our limited compute budget, we sample only 16 frames per video in our experiments.

\textbf{Category Distribution.}
The distribution over super-categories is visualized in \cref{fig:dataset_distr_super}. 
As shown, the dataset exhibits broad semantic coverage, spanning diverse domains from indoor to outdoor object categories, thereby capturing substantial real-world diversity. Importantly, the category distribution remains largely unchanged, with only minor deviations from the original dataset. These changes primarily affect categories such as animals, where dynamic videos break the assumption of a single consistent coordinate frame, and meals, which often contain multiple small objects.

\textbf{Viewpoint Distribution.}
\cref{fig:dataset_distr_viewpoint} illustrates the viewpoint distribution of the canonicalized objects. The dataset provides coverage from all viewing directions, indicating that objects are well represented across the full pose sphere. A slight bias toward frontal views and viewpoints mildly elevated above the horizontal plane is observable. This tendency can be attributed to common recording conditions in real-world videos: many objects naturally rest or are positioned upright (e.g., bicycles), or are placed such that their frontal surface faces upward (e.g., books). Consequently, front-facing and slightly top-down perspectives occur more frequently, while overall maintaining broad viewpoint diversity.

\textbf{Train--Test Split.}
Every9D-22M contains 960 categories in total. For each category, the test set contains at most five object instances, for which we select only the highest-quality annotations; all test labels were additionally double-checked by human annotators. The test set covers 898 categories, 630 of which contain five object instances; evaluation is conducted on all test categories. The number of categories and object instances per split is reported in \cref{tab:sup_split_statistics}. All remaining videos, covering 895 of the 960 categories, form the training set, which is not limited in the number of object instances per category and typically contains many more instances per category than the test set.

\textbf{Label Quality.}
Since the test set consists only of the highest-quality, double-checked annotations, the following quality assessment applies exclusively to the training set. To assess the accuracy of its generated 9D pose labels, we uniformly sample 125 approved training objects and independently annotate their poses by two human annotators; whenever the two annotations differ by more than $30^\circ$, the object is re-examined to establish the ground-truth orientation. Category-level symmetries are accounted for by accepting all equivalent orientations as valid.
We find that $92.8\%$ of the generated labels are accurate within $30^\circ$ and $85.6\%$ within $15^\circ$.
Of the 9 labels with errors exceeding $30^\circ$, 5 correspond to unavoidable $180^\circ$ ambiguities that cannot be resolved from appearance alone (e.g., a switched-off smartphone), 2 involve objects with an inherently ill-defined canonical orientation (e.g., a round hairbrush), and only 2 are genuine annotation mistakes, i.e., a false acceptance rate of merely $1.6\%$.
Errors between $15^\circ$ and $30^\circ$ predominantly involve objects lacking reliable geometric cues, such as curled cats, squeezed teddy bears, or nearly spherical fruits; notably, even the two independent annotators disagreed by more than $15^\circ$ in 4 of these 10 cases.
Overall, the small number of genuine annotation errors and the predominance of intrinsically ambiguous cases demonstrate the reliability of the generated labels.

\section{Experiments}
\label{sec:results}
We validate the key design choices underlying our canonicalization framework in \cref{tab:ablation}, and evaluate on the disjoint training and test splits of Every9D-22M (\cref{sec:dataset}).
As shown in \cref{tab:comparison}, training on Every9D-22M:
(a) generalizes better to other datasets than any other training dataset;
(b) outperforms prior foundation models in rotation accuracy on all benchmarks except ImageNet3D; and
(c) reveals a strong viewpoint bias of existing datasets, which our jointly trained model overcomes.
\subsection{Experimental Details}

\textbf{Datasets.}
We evaluate on all real-world pose datasets in~\cref{tab:datasets} with more than 15 categories, namely ImageNet3D~\citep{ma2024imagenet3d}, HANDAL~\citep{guo2023handal}, Omni6DPose-ROPE~\citep{zhang2024omni6dpose}, and Every9D-22M, except for ObjectNet3D~\citep{xiang2016objectnet3d}. ImageNet3D extends ObjectNet3D, aiming in particular to improve its annotation quality.

\textbf{Symmetry-Aware Evaluation.}
The cross-category orientation rules (\cref{sec:sup_orientations}) identify category-level symmetries through underconstrained orientations, as described in \cref{sec:method_pose_annotation}.
\begin{figure}[t]
    \includegraphics[width=0.97\linewidth]{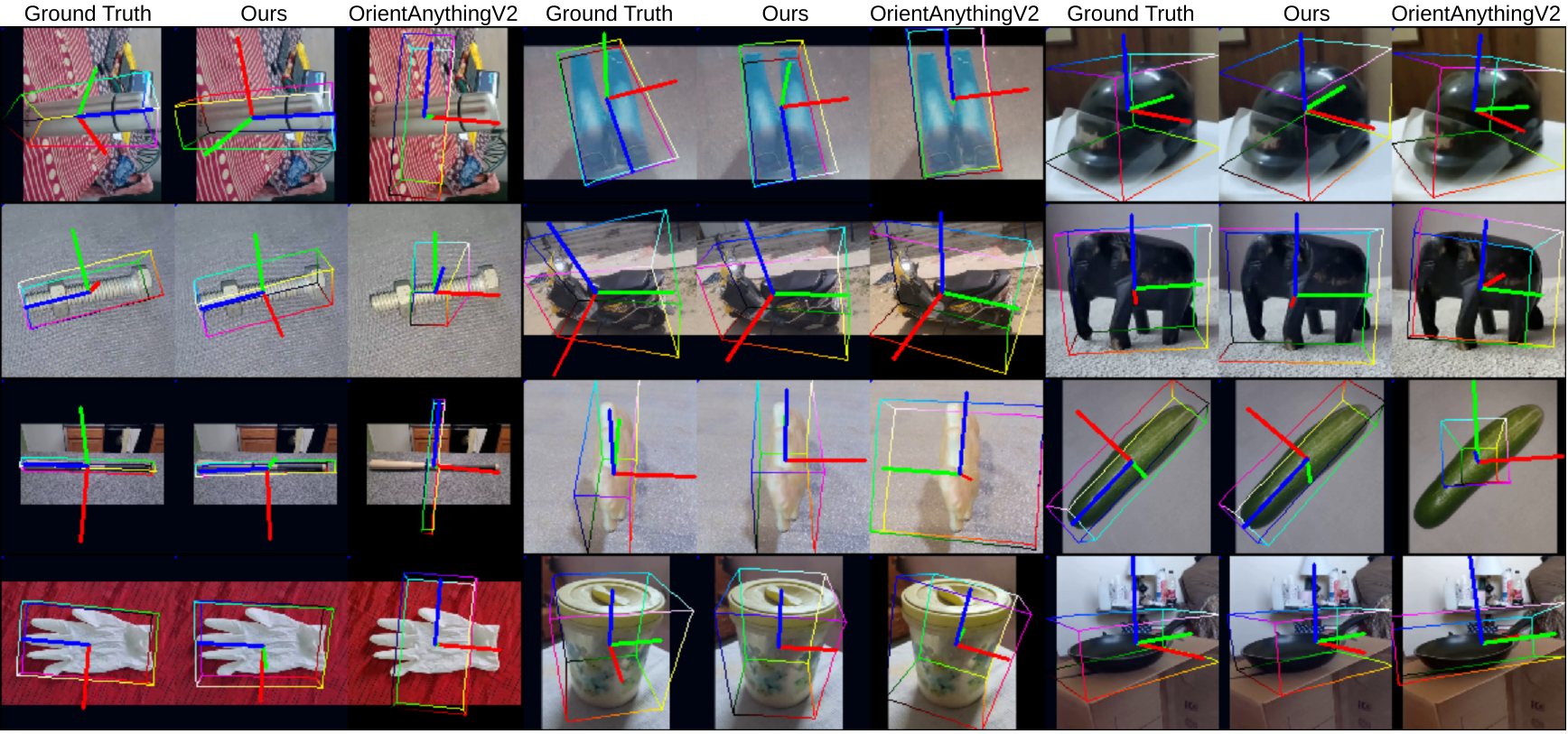}
    \caption{\textbf{Qualitative Results on Every9D-22M (test split).}
Since OrientAnythingV2 does not predict 3D bounding boxes, we visualize its rotation estimates using ground-truth 3D boxes. Predicted axes are color-coded as red (left), green (back), and blue (top).}
    \label{fig:results}
\end{figure}
\subsection{Comparison with Prior Work}
\textbf{Every9D-22M generalizes best across datasets.}
In the lower part of~\cref{tab:comparison}, we train our model on individual datasets and evaluate on all others.
On every benchmark other than its own, the model trained on Every9D-22M-train achieves the highest rotation accuracy among all training datasets.
Compared to the next best training dataset, it improves coarse ($30^\circ$) accuracy on ImageNet3D by \textbf{15.2} / \textbf{13.9} points (symmetry-unaware / symmetry-aware, vs.\ Omni6DPose-SOPE), on HANDAL by \textbf{3.9} / \textbf{3.3} points (vs.\ Omni6DPose-SOPE), and on Omni6DPose by \textbf{30.2} / \textbf{29.0} points (vs.\ ImageNet3D).
On Omni6DPose, it also improves 3D IoU by \textbf{25.9} points and, despite the domain gap, recovers most of the accuracy of training on Omni6DPose-SOPE (54.0 vs.\ 62.8).

\textbf{Every9D-22M outperforms prior foundation models.}
Every9D-22M poses a substantial challenge for state-of-the-art foundation models: the best of them reaches only 23.8 / 47.4 coarse ($30^\circ$) accuracy on Every9D-22M-test.
Our model trained on Every9D-22M-train outperforms the best foundation model in coarse rotation accuracy by \textbf{14.4} / \textbf{14.0} points on Every9D-22M-test, by \textbf{8.6} / \textbf{21.0} points on HANDAL, and by \textbf{18.7} / \textbf{18.8} points on Omni6DPose (symmetry-unaware / symmetry-aware).
ImageNet3D is the only exception, discussed below.
The qualitative comparisons in~\cref{fig:results} highlight three key advantages of our approach over OrientAnythingV2: improved robustness to unusual object orientations, more accurate estimation of subtle angular differences, and stronger resilience under challenging lighting conditions.
Furthermore, WildDet3D orientations appear strongly biased by gravity, resulting in poor rotation accuracy.
In terms of 3D IoU, our model outperforms WildDet3D by \textbf{28.0} points and SAM3D by \textbf{9.8} points on Every9D-22M-test.

\textbf{Existing datasets exhibit a strong viewpoint bias.}
Models trained on a single dataset perform well in-domain but degrade on other datasets, most pronouncedly for ImageNet3D, whose viewpoints concentrate sharply along the equator (\cref{fig:dataset_distr_viewpoint}).
Our model trained on ImageNet3D achieves 58.6 / 64.2 coarse accuracy in-domain, but only 4.2 / 10.0 on HANDAL and 10.2 / 28.0 on Every9D-22M-test; OrientAnythingV2, whose training data includes ImageNet3D, shows the same pattern (2.1 / 3.8 on HANDAL).
Conversely, models trained on broader viewpoint coverage perform considerably worse on ImageNet3D, which thus rewards a prior towards canonical photograph viewpoints.
To leverage both, we jointly train our model with mixed batches, where half of each batch is sampled from Every9D-22M-train and half from ImageNet3D (\cref{sec:sup_training_details}).
The jointly trained model outperforms all prior foundation models on ImageNet3D, improving over OrientAnythingV2 by \textbf{3.4} / \textbf{2.7} points, and slightly improves over training on ImageNet3D alone (59.1 / 65.4 vs.\ 58.6 / 64.2), while substantially improving over it on all other benchmarks (\cref{tab:comparison}).

\begin{table*}[t]
\centering
\small
\setlength{\tabcolsep}{4pt}
\renewcommand{\arraystretch}{1.15}

\resizebox{\linewidth}{!}{%
\begin{tabular}{lccccccccccc}
\toprule

Dataset
& \multicolumn{2}{c}{ImageNet3D}
& \multicolumn{3}{c}{HANDAL}
& \multicolumn{3}{c}{Omni6DPose-ROPE}
& \multicolumn{3}{c}{Every9D-test} \\

\cmidrule(lr){2-3}
\cmidrule(lr){4-6}
\cmidrule(lr){7-9}
\cmidrule(lr){10-12}

& \multicolumn{2}{c}{Acc@$30^\circ$}
& \multicolumn{2}{c}{Acc@$30^\circ$}
& 3D IoU
& \multicolumn{2}{c}{Acc@$30^\circ$}
& 3D IoU
& \multicolumn{2}{c}{Acc@$30^\circ$}
& 3D IoU \\

\cmidrule(lr){2-3}
\cmidrule(lr){4-5}
\cmidrule(lr){7-8}
\cmidrule(lr){10-11}

Symmetry-Aware
& $\times$ & \checkmark
& $\times$ & \checkmark
&
& \checkmark$^*$ & \checkmark
&
& $\times$ & \checkmark
& \\

\midrule

OrientAnythingV1~\citep{wang2024orient}
& 41.6 & 48.4
& 2.9 & 7.4
& --
& 27.7 & 29.8
& --
& 18.6 & 37.5
& -- \\

OrientAnythingV2$^\ddagger$~\citep{wang2026orient}
& \underline{55.7} & \underline{62.7}
& 2.1 & 3.8
& --
& 31.2 & 33.4
& --
& \underline{23.8} & 46.1
& -- \\

WildDet3D (RGB-D)~\citep{huang2026wilddet3d}
& 27.4 & 37.5
& 1.3 & 2.7
& 7.0
& 25.1 & 26.9
& 4.3
& 12.4 & 33.5
& 22.6 \\

SAM3D (RGB-D)~\citep{chen2025sam}
& 46.3 & 55.0
& \underline{6.7} & \underline{8.8}
& \textbf{29.3}
& \underline{35.3} & \underline{35.3}
& \underline{39.5}
& 23.6 & \underline{47.4}
& \underline{40.8} \\

Ours (ImageNet3D + Every9D)
& \textbf{59.1} & \textbf{65.4}
& \textbf{12.2} & \textbf{23.2}
& \underline{23.0}
& \textbf{51.8} & \textbf{51.4}
& \textbf{39.6}
& \textbf{33.7} & \textbf{57.7}
& \textbf{48.0} \\

\midrule

Ours (OrientAnythingV2$^\dagger$)
& 15.2 & 24.6
& 0.7 & 2.6
& 13.9
& 11.9 & 13.9
& 10.2
& 3.2 & 13.7
& 9.0 \\

Ours (ImageNet3D)
& \textbf{58.6} & \textbf{64.2}
& 4.2 & 10.0
& 15.4
& 23.8 & 25.1
& 10.7
& 10.2 & 28.0
& 9.6 \\

Ours (HANDAL)
& 3.4 & 7.1
& \textbf{64.7} & \textbf{77.7}
& \textbf{46.4}
& 15.7 & 17.1
& 15.9
& 4.6 & 17.1
& 22.3 \\

Ours (Omni6DPose-SOPE$^\dagger$)
& 15.7 & 25.3
& 11.4 & 26.5
& \underline{27.5}
& \textbf{62.8} & \textbf{61.1}
& \textbf{45.2}
& \underline{22.8} & \underline{46.9}
& \underline{40.7} \\

Ours (Every9D)
& \underline{30.9} & \underline{39.2}
& \underline{15.3} & \underline{29.8}
& 26.9
& \underline{54.0} & \underline{54.1}
& \underline{41.8}
& \textbf{38.2} & \textbf{61.4}
& \textbf{50.6} \\

\bottomrule
\end{tabular}%
}

\caption{
\textbf{Quantitative Results Across Datasets.}
We report rotational accuracy at the coarse ($30^\circ$) threshold, without and with symmetry-aware evaluation, and the average 3D IoU. The upper block compares our approach, trained jointly on ImageNet3D and Every9D, against foundation models; the lower block reports our approach trained on individual datasets. Best and second-best results within each block are \textbf{bold} and \underline{underlined}. Dashes mark rotation-only models.
$^*$For Omni6DPose, this column is evaluated with the dataset's instance-level symmetries instead of our category-level ones.
$^\ddagger$OrientAnythingV2 is also trained on ImageNet3D.
$^\dagger$Synthetic training datasets.}
\label{tab:comparison}

\end{table*}

\section{Limitations}
\textbf{Symmetry handling at training time.}
Our evaluation accounts for category-level symmetries, but the baseline model itself predicts a single deterministic pose per image. Modeling symmetry-induced ambiguity through multi-hypothesis or distributional predictions is a promising direction for improving performance on rotationally symmetric categories.

\textbf{Depth and dynamics.}
Depth annotations are inherited from monocular DepthAnythingV2~\citep{yang2024depth} estimates rather than ground-truth sensor depth, with scale aligned to the structure-from-motion reconstruction. The dataset also does not cover dynamic object interactions: each video is assumed to capture a static instance from multiple viewpoints.
\section{Conclusion}

We introduced \textbf{Every9D-22M}, 9D pose annotations for 22.8M real-world images across 630 everyday object categories, canonicalized from 114K object-centric videos with full human verification of every pose. Training on it generalizes better than on existing datasets, and joint training with ImageNet3D surpasses all prior foundation models in rotation accuracy on every benchmark.

\section*{AI use statement}

We used generative AI tools as general-purpose writing and coding assistants.
Specifically, they were used to polish the phrasing of text that we wrote
ourselves, and to assist with routine implementation tasks such as boilerplate
code, plotting scripts, and refactoring. We did not use generative AI tools to
generate research ideas, to derive or verify any of the claims made in this
paper, to produce or label any part of Every9D-22M, or to retrieve and
summarize related work. All AI-assisted text and code was reviewed by the
authors; AI-assisted code was verified by inspection and by the experiments
reported in \cref{sec:results} and the supplementary material. We take
responsibility for the final content of this work, including all text, claims,
and artifacts.

\section*{Ethics statement}

Every9D-22M is built entirely on top of uCO3D~\citep{liu2025uncommon}, which we
use in accordance with its official license terms (CC BY 4.0); we credit the
original authors accordingly. The underlying videos are object-centric recordings
of everyday objects rather than human-centered data, so the dataset contains no
personally identifying subjects by construction, and we introduce no new data
collection involving human subjects. The manual annotation and verification
described in \cref{sec:sup_manual_labor} was performed by the authors, not by
crowdworkers. We discuss the broader risks of large-scale 9D pose estimation,
including surveillance misuse, category and geographic imbalance in the source
videos, and the propagation of monocular depth errors into downstream systems,
in \cref{sec:sup_societal_impact}.

\section*{Reproducibility statement}

The canonicalization framework is described in \cref{sec:method}, with the
design choices behind it ablated in \cref{sec:sup_design_choices}. The dataset
construction, filtering criteria, and train/test splits are given in
\cref{sec:dataset} and \cref{sec:sup_dataset_details}, and the symmetry-aware
evaluation protocol together with the orientation rules for every category is
specified in \cref{sec:sup_orientations}. Full training details for the
baseline models, including losses and weights, point cloud
normalization, optimizer, schedule, batch size, and parameter counts, are
reported in \cref{sec:sup_training_details}. The annotation and verification
interfaces and the human effort they required are documented in
\cref{sec:sup_manual_labor}. Data and code will be public upon acceptance.

\ificlrfinal
\section*{Acknowledgments}
AK acknowledges support via his Emmy Noether Research Group funded by the German Research Foundation (DFG) under grant number 468670075.
This research was funded by the Deutsche Forschungsgemeinschaft (DFG, German Research Foundation) under grant number 539134284, through EFRE (FEIH\_2698644) and the state of Baden-Württemberg. 
\begin{center}
\includegraphics[width=0.3\textwidth]{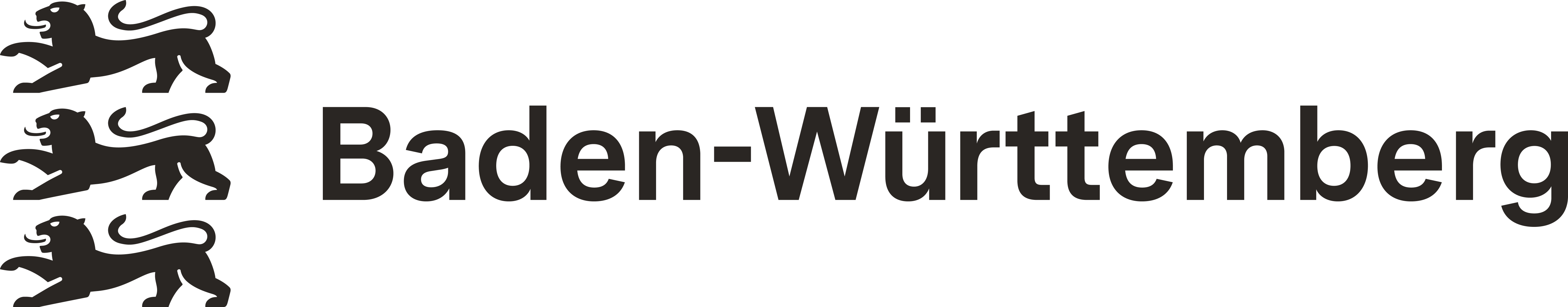} ~~~ \includegraphics[width=0.3\textwidth]{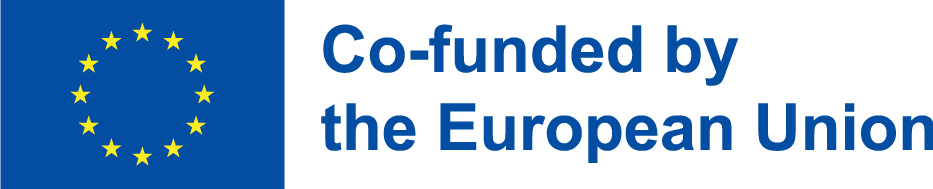} 
\end{center}
\fi

\bibliographystyle{iclr2027_conference}
\bibliography{main}

\newpage

\begin{center}
    {\LARGE \textbf{Supplementary Material}}\\[1em]
    {\large Every9D-22M: Large-Scale Real-World 9D Canonicalization of Everyday Objects}
\end{center}

\appendix

\renewcommand{\thesection}{S\arabic{section}}
\renewcommand{\thefigure}{S\arabic{figure}}
\renewcommand{\thetable}{S\arabic{table}}

\renewcommand{\theHfigure}{S\arabic{figure}}
\renewcommand{\theHtable}{S\arabic{table}}

\setcounter{section}{0}
\setcounter{figure}{0}
\setcounter{table}{0}

This supplementary material provides additional details and results that complement the main paper. In \cref{sec:sup_design_choices}, we present further ablations on the design choices that enabled our method to scale to more than 100K objects with limited computational resources. Further, in \cref{sec:sup_dataset_details}, we provide more details about the dataset and in \cref{sec:sup_training_details} about the training. In \cref{sec:sup_orientations}, we describe how orientations are defined for all LVIS categories appearing in uCO3D~\citep{liu2025uncommon}. In \cref{sec:sup_manual_labor}, we show our annotation and verification tools and quantify the human effort required to ensure a high-quality dataset. In \cref{sec:sup_qualitative_examples}, we provide additional qualitative examples of our canonicalized objects, and finally we discuss societal impact in \cref{sec:sup_societal_impact}.

\section{Canonicalization Design Choices}
\label{sec:sup_design_choices}

In \cref{tab:sup_abl_design_backbone_frames}, we illustrate that CLIP features perform worse than DINOv3 features, likely because the contrastive objective separates semantically related categories (e.g., motorcycles and bicycles) despite their similar geometry. 
Furthermore, using 32 frames performs only 1.5 percentage points worse than using 64 frames, while substantially reducing computation. This efficiency enables our approach to scale to over 100K objects.

In~\cref{tab:verification_statistics}, we quantify the verification process. Overall, $74.9\%$ of videos are successfully annotated, while the remaining ones are mostly unsuitable for pose estimation due to poor reconstruction or the presence of multiple objects.

\section{Dataset details.}
\label{sec:sup_dataset_details}
\begin{figure}[t]
  \centering
  \includegraphics[width=\textwidth]{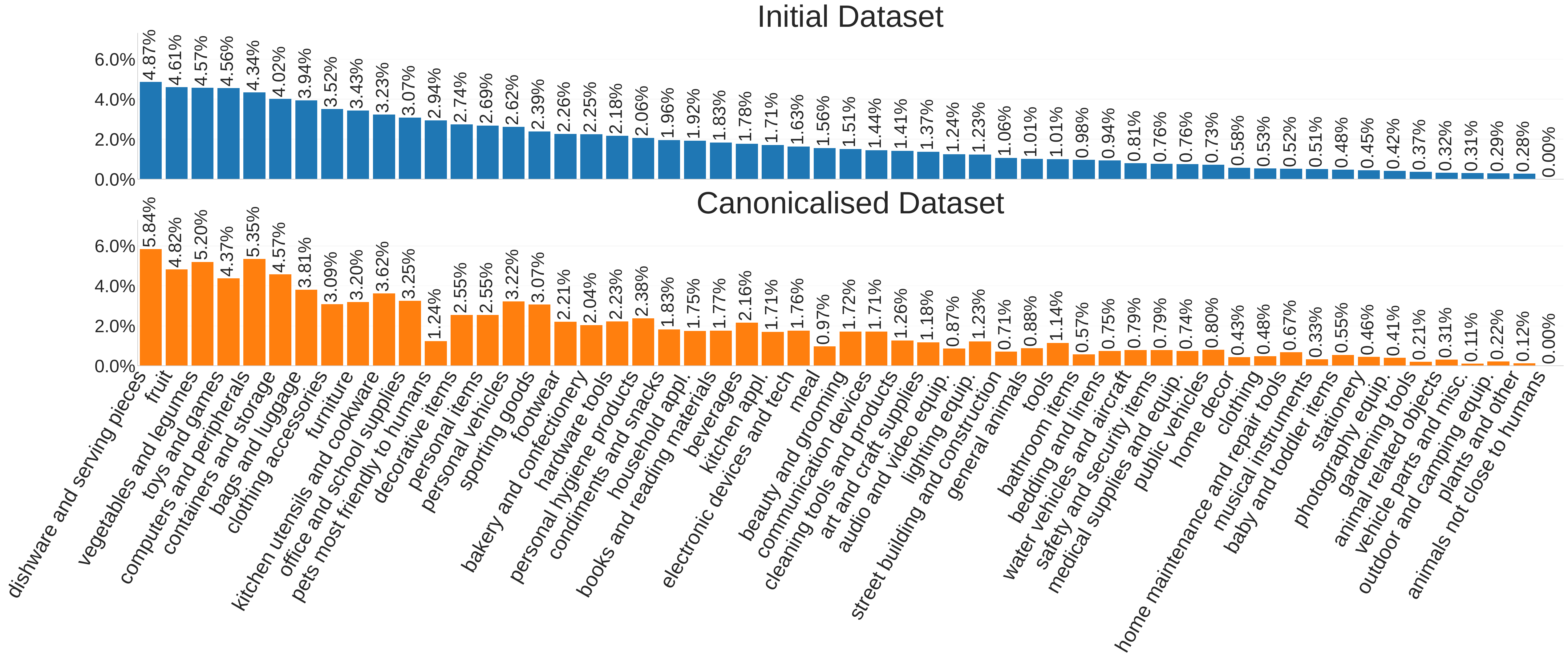}
  \caption{\textbf{Super-Category Distribution.} Every9D-22M covers a broad range of object categories. Our filtering process preserves the broad distribution.}
  \label{fig:dataset_distr_super}
\end{figure}

We provide the category distribution across super-categories in \cref{fig:dataset_distr_super}.

For each category, the test split contains at most five object instances (videos), for which we select only the highest-quality annotations, resulting in a balanced set of test videos. \cref{tab:sup_split_statistics} reports the number of categories and object instances per split; evaluation is conducted on all 898 test categories, 630 of which contain five test instances. To obtain a computationally manageable yet diverse evaluation set, we uniformly sample 16 frames from each video. Each image is then cropped using the bounding box derived from its corresponding object mask.

All remaining videos are used for training; in contrast to the test split, the number of object instances per category in the training split is not limited. While evaluation is conducted on the 898 test categories, the training set covers 895 of the 960 categories. Each evaluated category is covered by at least as many training object instances as its number of test object instances minus one.

\section{Training Details.} 
\label{sec:sup_training_details} 
We weight the translation, rotation, and object size loss with $1.$, $0.1$, and $1.$.

We observed instability when training LitePT directly on the raw input point clouds. To mitigate this issue, we normalize each point cloud by subtracting the median and scaling using the interquartile range. Additionally, we clamp points beyond a predefined maximum distance in the normalized space.

We train our model for 100K iterations with a global batch size of 128. Training is conducted on 4 GPUs, where each batch is evenly distributed across devices. At the beginning of every epoch, we perform both inter-shard shuffling and intra-shard shuffling to ensure effective data randomization and reduce sampling bias.

\textbf{Joint training with ImageNet3D.} For ImageNet3D samples, we obtain segmentation masks using SAM~\citep{kirillov2023segment}, prompted with the 2D bounding boxes, and estimate depth with DepthAnythingV3~\citep{lin2025depth}. Since the scene scale is not aligned with the predicted depth, we only calculate the orientation loss for these samples.

Following \citet{zhang2025beyond}, we adopt the AdamW optimizer~\citep{loshchilov2017decoupled} with an initial learning rate of $1\times10^{-4}$ and a weight decay of $5\times10^{-4}$. We employ an exponential learning rate scheduler with a decay factor of $0.98$. Our model comprises 37M parameters, of which 15M are trainable.

\begin{table}
\centering
\caption{\textbf{Train--Test Split Statistics.} Number of categories with more than $k$ object instances per split, and total number of object instances. Evaluation is conducted on all test categories.}
\label{tab:sup_split_statistics}
\begin{tabular}{lrrr}
\toprule
Categories with & Train & Test & Train + Test \\
\midrule
$>0$ instances & 895 & 898 & 960 \\
$>1$ instances & 774 & 766 & 856 \\
$>2$ instances & 706 & 700 & 789 \\
$>3$ instances & 667 & 657 & 753 \\
$>4$ instances & 639 & 630 & 721 \\
$>5$ instances & 621 & 0   & 693 \\
\midrule
Total instances & 110,882 & 3,651 & 114,533 \\
\bottomrule
\end{tabular}
\end{table}

\begin{table}
\centering
\caption{\textbf{Canonicalization Design Choices.}
We evaluate key design decisions for clustering and automatic alignment on the annotated subset of CO3D~\citep{reizenstein2021common,goodwin2022zero}.}
\label{tab:sup_abl_design_backbone_frames}
\begin{tabular}{c c c}
\toprule
Backbone & \#Frames & Acc@30$^\circ$ \\
\midrule
DINOv3-ViT-L & 32 & 82.2 \\
DINOv3-ViT-L & 64 & +1.5 \\
DINOv3-ViT-L & 16 & -2.7 \\
DINOv3-ViT-L & 4  & -3.4 \\
CLIP-ViT-L   & 64 & -5.4 \\
\bottomrule
\end{tabular}
\end{table}

\begin{table}
\centering
\caption{\textbf{Verification Statistics.}}
\label{tab:verification_statistics}
\footnotesize
\setlength{\tabcolsep}{3pt}
\begin{tabular}{lrrr}
\toprule
Iter. & Accept & Skipped & Filtered \\
\midrule
1st  & 63.3\% & 19.6\% & 17.1\% \\
2nd  & 59.0\% & 35.8\% & 5.2\%  \\
\midrule
Total & 74.9\% & 7.0\% & 18.1\% \\
\bottomrule
\end{tabular}
\end{table}

\section{Orientations}
\label{sec:sup_orientations}

We define three categories of directions: 24 intrinsic object directions, 17 extrinsic object–user directions, and 5 extrinsic object–user–object directions. Note that some relations correspond to axes rather than directed orientations due to underlying symmetries.

\subsection{Intrinsic directions.}

\noindent "move backward": Opposite direction of natural forward movement, e.g. cow.

\noindent "move upward": Upward direction of movement or during movement, e.g. frisbee. 

\noindent "move backward and forward": Axis of natural forward and backward movement, e.g. skateboard.

\noindent "move upward and downward": Axis of upward and downward during movement, e.g. kite.

\noindent "contain or support upward": Direction upwards of supporting or containing other objects or liquids. e.g. bag. 

\noindent "contain or support upward and downward": Axis of supporting or containing other objects or liquids, e.g. hourglass.

\noindent "cover upward": Direction of covering from falling objects or liquid, e.g. umbrella.

\noindent "cover upward and downward": Axis of covering from falling objects or liquid, e.g. doormat.

\noindent "entry backward": Opposite direction of entry, e.g. kennel. 

\noindent "stand upward": Natural standing direction. Often defined via gravity.

\noindent "stand upward downward": Natural standing axis. Often defined via gravity.

\noindent "face backward": Backward of face for things with face that are not moving, e.g. teddy bears. 

\noindent "sprout upward": Direction of sprout, e.g. onion. 

\noindent "sprout upward and downward": Axis of sprout, when direction is unclear, e.g. log.

\noindent "stem upward": Stem direction of hanging fruits, e.g. apples. 

\noindent "stem upward and downward": Axis of stem, when direction is unclear, e.g. lemon.

\noindent "winding axis": Winding axis for e.g. tape.

\noindent "flow backward": Opposite of flow direction, flow may be arrows, liquid, wind, light, or electric signal.

\noindent "flow backward and forward": Axis of flow direction.   

\noindent "unwinding extension backward": Opposite direction of unwinding. 

\noindent "eyelet axis": Axis of eyelet, e.g. wristband, or, rubberband.

\noindent "toward mount backward": Backward horizontal mounting direction, e.g. sink.

\noindent "toward and away from mount backward": Mounting horizontal axis, e.g. deadbolt.  

\noindent "toward mount upward":  Upward vertical mounting direction, e.g. lightbulb.

\subsection{Extrinsic object-user directions.}

\noindent "away from user": Direction that points away from user, e.g faucet. 

\noindent "away from user fingers": Direction that points away from user fingers, e.g. keyboard. 

\noindent "toward and away from user": Axis that points towards and away from user, e.g. ironing board.

\noindent "user upward": Direction of user upward, e.g. raincoat.

\noindent "user backward": Direction of user backward, e.g. raincoat.

\noindent "user upward and downward": Axis of user upward and downward, e.g. headband.

\noindent "user lying axis head feet": Axis of user lying between head and feet, e.g. bed. 

\noindent "user forward and backward": Axis of user forward and backward, e.g. headset.

\noindent "along grab two hands left right": Axis of hands grabbing left and right, e.g. tray.

\noindent "along grab two hands away from function": Opposite of functional direction for tools required two hands, e.g. mop. %

\noindent "along grab one hand": Axis for symmetric functional directions of tools grabbed with one hand, e.g. wrench. %

\noindent "along grab one hand toward function": Functional direction for tools grabbed with one hand, e.g. hammer.

\noindent "along stand grab one hand grab away from function": Opposite of functional direction for tools which can stand, but also be grabbed with one hand, e.g. pan.

\noindent "along grab two fingers toward function": Functional direction for tools grabbed with two fingers, e.g. pencil. 

\noindent "hand fingers": Direction of user's finger tip, e.g. glove. %

\noindent "hand back": Direction of user's hand back, e.g. baseball glove.

\noindent "hand backward and forward": Axis of user's hand back and palm, e.g. glove.

\subsection{Extrinsic object-user-object directions.}

\noindent "away from object": Direction pointing away from another object, e.g. spoon.

\noindent "toward and away from object": Direction of point away and toward another object, e.g. racket. 

\noindent "object upward": Direction of upward of another object, e.g. corkscrew. 

\noindent "object squeeze": Axis of squeezing an object, e.g. scissor. 

\noindent "object squeeze back": Direction toward the squeezing side that is moving more, whereas the other part is usually always in contact with the other object, e.g. barrette. 

\subsection{Category-Level Mapping.}
\label{sec:sup_orientation_mapping}

We provide category-level mappings to our defined orientations for 1076 LVIS categories that appear in uCO3D~\citep{liu2025uncommon}. Note that we can reduce the 1076 categories to 101 orientation categories.

\noindent LEFT -, BACK user backward, TOP stand upward: bench, chair, sofa, slide, armchair, chaise longue, rocking chair, recliner, folding chair, loveseat, cabinet, sofa bed, deck chair, mouse computer equipment, vacuum cleaner, bedpan, highchair, grinder, treadmill

\noindent LEFT -, BACK move backward, TOP move upward: airplane, dog, cart, lawn mower, bicycle, tricycle, school bus, race car, cab taxi, jeep, convertible automobile, motorcycle, dirt bike, golfcart, pickup truck, bus vehicle, motor vehicle, car automobile, motor scooter, minivan, fire engine, giraffe, cock, ram animal, chicken animal, turtle, penguin, deer, hog, bird, mallard, owl, monkey, camel, cow, frog, horse, seahorse, crow, zebra, goat, duckling, goose, black sheep, domestic ass, elephant, sheep, giant panda, duck, flamingo, dove, gazelle, cub animal, polar bear, seabird, elk, bull, lamb animal, alligator, gorilla, crab animal, calf, parrot, foal, pony, dolphin, rabbit, bulldog, squirrel, puppy, shepherd dog, goldfish, kitten, parakeet, cat, dalmatian, hamster, ferret, pet, rodent, pug dog, handcart, grizzly, prawn, barrow, plow farm equipment, ferry, cruise ship, jet plane, cargo ship, boat, water scooter, passenger ship, blimp, helicopter, river boat, fighter jet, yacht, space shuttle, train railroad vehicle, garbage truck, trailer truck, truck, tractor farm equipment, police cruiser, bullet train, army tank, forklift, tow truck, bulldozer, ambulance, horse buggy, camper vehicle, wheelchair, baby buggy, toy

\noindent LEFT -, BACK toward and away from user, TOP contain or support upward: handbag, suitcase, briefcase, envelope, box, mat gym equipment, wristlet, identity card, shoulder bag, pouch, shopping bag, tote bag, cooler for food, duffel bag, ironing board, pencil box

\noindent LEFT -, BACK away from user, TOP contain or support upward: backpack, palette, coloring material, trash can, locker, bread bin, mailbox at home, perfume, cigar box, saddlebag, clutch bag, wallet, satchel, cigarette case, hatbox, toolbox, dumpster, towel rack, water cooler

\noindent LEFT -, BACK away from user, TOP along grab one hand toward function: hair dryer, razorblade, hairbrush, shaver electric, toothbrush

\noindent LEFT -, BACK hand back, TOP hand fingers: baseball glove, boxing glove

\noindent LEFT -, BACK toward and away from object, TOP along grab one hand toward function: tennis racket, racket, bottle opener, peeler tool for fruit and vegetables, spatula, file tool, scraper, hammer, mallet, sword, dagger

\noindent LEFT -, BACK -, TOP along grab one hand toward function: baseball bat, igniter, blender, crossbar, sparkler fireworks, eggbeater, skewer, cooking utensil, stirrer, masher, chap, boom microphone, microphone, flashlight, screwdriver, drumstick, whisk

\noindent LEFT -, BACK move backward and forward, TOP move upward: skateboard, canoe, raft, houseboat, drone, railcar part of a train, cabin car, freight car, wagon, passenger car part of a train

\noindent LEFT -, BACK -, TOP -: ball, cube, ping pong ball, volleyball, softball, soccer ball, gemstone, baseball, basketball, tennis ball, bowling ball, leather, die, beachball, sherbert, batter food, salsa, crouton, cast, potato

\noindent LEFT -, BACK -, TOP along grab one hand: sausage, dumbbell, chocolate bar, string cheese, candy bar, egg roll, eraser, eclair

\noindent LEFT -, BACK -, TOP stand upward: sandwich, magnet, pad, paperweight, lightning rod, windsock, solar array, manhole, silo, water tower, streetlight, pole, baseball base, vase, candle holder, wreath, candle, flowerpot, pizza, cake, lego, checkerboard, gameboard, pool table, bun, brownie, cornbread, muffin, french toast, bread, pita bread, christmas tree, beanbag, dining table, footstool, coffee table, ottoman, table, kitchen table, stool, lamppost, lamp, oil lamp, lantern, birdbath, sail, cabana, tortilla, funnel, music stool, drum musical instrument, table lamp, chocolate cake, quiche, dish, casserole, battery, cone, doughnut, piggy bank, door, yurt, urn, ashtray, beef food, salmon food, hamburger, octopus food, steak food, bagel, turkey food, egg yolk, crabmeat, tripod, playpen, waffle, pretzel, crumb, truffle chocolate, toast food, pie, birthday cake, cookie, pastry, cupcake, bowling pin

\noindent LEFT -, BACK user backward, TOP user upward: camera, costume, raincoat, vest, overalls clothing, blazer, pajamas, suit clothing, bridal gown, trench coat, jean, corset, flannel, nightshirt, fleece, kilt, sportswear, sweater, parka, shirt, cardigan, jacket, tux, turtleneck clothing, ballet skirt, dress, coat, jersey, kimono, robe, lab coat, swimsuit, halter top, shoe, goggles, football helmet, rollerblade, helmet, choker, turban, bow tie, cowboy hat, bandanna, spectacles, neckerchief, tiara, ice skate, roller skate, eyepatch, breechcloth, tartan, poncho, blinder for horses, bullhorn, camcorder, joystick, whistle, headscarf, baseball cap, bolo tie, lanyard, necklace, brassiere, tights clothing, sunglasses, crown, sock, visor, necktie, beret, apron, fedora, ski parka, cap headwear, cloak, gasmask, crutch, cornet, saxophone, bagpipe, bass horn, clarinet, noseband for animals, nosebag for animals, armor, chain mail, mask, bulletproof vest, slipper footwear, boot, flip flop sandal, sandal type of shoe, ski boot, flipper footwear, arctic type of shoe, wooden leg, television camera, binoculars, bib, musical instrument

\noindent LEFT -, BACK -, TOP sprout upward: carrot, garlic, brussels sprouts, onion, cauliflower, asparagus, edible corn, ginger, green onion, tomato, green bean, legume, celery, turnip, mushroom, broccoli, radish, sweet potato, flower arrangement, carnation, bamboo, sunflower, bouquet, artichoke, lettuce

\noindent LEFT -, BACK -, TOP stem upward: cucumber, pepper, eggplant, gourd, grape, persimmon, clementine, mandarin orange, peach, blueberry, date fruit, apple, pineapple, cantaloup, pear, blackberry, orange fruit, almond, pinecone, nut, raspberry, prune, coconut, apricot, avocado, papaya, fig fruit, strawberry, cherry, bell pepper, pumpkin, chili vegetable, zucchini, chickpea

\noindent LEFT -, BACK away from user, TOP stand upward: parking meter, stop sign, fireplug, easel, sewing machine, shredder for paper, toaster, microwave oven, refrigerator, dishwasher, waffle iron, toaster oven, oven, coffee maker, mixer kitchen tool, food processor, stove, grill, postbox public, vending machine, telephone booth, elevator car, traffic light, street sign, newsstand, trunk, milestone, figurine, billboard, basketball backboard, window box for plants, subwoofer, cd player, radio receiver, amplifier, speaker stero equipment, stereo sound system, record player, phonograph record, car battery, dollhouse, award, wardrobe, armoire, dresser, drawer, desk, cupboard, cash register, file cabinet, television set, laptop computer, piano, harmonium, microscope, globe, printer, router computer equipment, gargoyle, hookah, heater, radiator, fan, water heater, automatic washer, air conditioner, hotplate, generator, ice maker, first aid kit

\noindent LEFT along grab two hands left right, BACK toward and away from user, TOP contain or support upward: grocery bag, plastic bag

\noindent LEFT -, BACK -, TOP along grab two fingers toward function: needle, knitting needle, pencil, thumbtack, pen, marker, cigarette, stylus, bolt, crayon, sharpener, sharpie, pencil sharpener, dropper, syringe, lollipop

\noindent LEFT -, BACK -, TOP winding axis: bobbin, thread, cylinder, garden hose, toilet tissue, dental floss, tape sticky cloth or paper, hose, bandage, coil, tinfoil, paper towel, sushi, sling bandage

\noindent LEFT -, BACK toward and away from user, TOP along grab one hand toward function: paintbrush

\noindent LEFT -, BACK -, TOP hand fingers: thimble

\noindent LEFT -, BACK -, TOP eyelet axis: bead, wristband, bracelet, wedding ring, armband, anklet, rubber band, pipe, straw for drinking, ring, earring

\noindent LEFT -, BACK object squeeze, TOP along grab one hand toward function: pliers, shears, hair curler, nutcracker, tongs, scissors, curling iron

\noindent LEFT -, BACK away from object, TOP along grab one hand toward function: ax, handsaw, fork, power shovel, kettle, spoon, strainer, steak knife, wooden spoon, ladle, pitchfork, knife, shovel, pocketknife, soupspoon

\noindent LEFT -, BACK toward and away from user, TOP stand upward: step stool, ladder, manger, table tennis table, clock tower, futon, stepladder, crescent roll, baguet, trophy cup, bicycle bump

\noindent LEFT -, BACK flow backward, TOP stand upward: reamer juicer, dish antenna, machine gun, fire extinguisher, cream pitcher, glass drink container, projector, webcam

\noindent LEFT -, BACK away from user, TOP user upward: calendar, clipboard, binder, scoreboard, painting, passport, tachometer, book, rearview mirror, cellular telephone, earphone, telephone, birthday card, booklet, paperback book, map, hardback book, diary, phonebook, comic book, postcard, bible, road map, mirror, clock, inkpad, notebook, checkbook, inhaler, control, pocket watch, timer, watch, remote control, calculator, ipod, beeper, alarm clock, water faucet, faucet

\noindent LEFT -, BACK object squeeze back, TOP along grab one hand toward function: stapler stapling machine, can opener, clippers for plants

\noindent LEFT -, BACK away from object, TOP stand upward: puncher, iron for clothing

\noindent LEFT -, BACK object squeeze, TOP along grab two fingers toward function: clip, hairpin, bobby pin, pin non jewelry, clothespin

\noindent LEFT -, BACK -, TOP contain or support upward: jar, barrel, beer bottle, garbage, milk can, thermos bottle, beer can, lemonade, orange juice, chocolate milk, alcohol, tequila, smoothie, soya milk, fruit juice, milk, liquor, vodka, wine bottle, pop soda, milkshake, cider, root beer, martini, hot sauce, cocoa beverage, pepper mill, colander, wineglass, bowl, watering can, toothpick, spice rack, shaving cream, toothpaste, gargle, lip balm, bottle, paper plate, flute glass, salad plate, vat, pottery, gift wrap, packet, tank storage vessel, water bottle, clothes hamper, bucket, canister, birdfeeder, cleansing agent, hand glass, detergent, dishwasher detergent, atomizer, hamper, can, carton, crate, water jug, chocolate mousse, grits, whipped cream, honey, yogurt, peanut butter, hummus, bubble gum, vinegar, cayenne spice, cornmeal, icecream, jam, tabasco sauce, fudge, olive oil, sour cream, soup, mint candy, cracker, jelly bean, condiment, applesauce, fishbowl, nest, aquarium, soup bowl, place mat, sugar bowl, saucer, saltshaker, shot glass, chinaware, dixie cup, wine bucket, shaker, plate, aerosol can, bathtub, platter, chalice, squid food, crisp potato chip, salad, lamb chop, ham, stew, meatball, patty food, lasagna, coleslaw, omelet, bean curd, mashed potato, rib food, pudding, pickle, gelatin, cistern, shampoo, pot, dispenser, freshener, medicine

\noindent LEFT along grab two hands left right, BACK -, TOP contain or support upward: keg, wok, basket, tray

\noindent LEFT -, BACK unwinding extension backward, TOP winding axis: tape measure, duct tape

\noindent LEFT -, BACK along stand grab one hand grab away from function, TOP stand upward: cappuccino, cooker, frying pan, pan for cooking, pan metal container, saucepan, cup, mug, dustpan, teacup

\noindent LEFT along grab two hands left right, BACK away from user, TOP stand upward: crock pot

\noindent LEFT -, BACK flow backward, TOP contain or support upward: teakettle, coffeepot, measuring cup, gravy boat, pitcher vessel for liquid, teapot

\noindent LEFT -, BACK toward and away from user, TOP user upward: crucifix, saddle blanket, dollar, money, cassette, videotape, penny coin, tag, business card, parchment, card, bookmark, notepad, keycard, air hammer

\noindent LEFT -, BACK toward mount backward, TOP cover upward: awning

\noindent LEFT -, BACK move backward, TOP contain or support upward: shopping cart

\noindent LEFT -, BACK toward mount backward, TOP stand upward: windmill, fireplace, pinwheel

\noindent LEFT -, BACK -, TOP sprout upward and downward: log, sugarcane plant

\noindent LEFT -, BACK flow backward and forward, TOP stand upward: telephone pole, mast

\noindent LEFT -, BACK hand back, TOP along grab one hand: football american

\noindent LEFT -, BACK flow backward, TOP user upward: bow weapon

\noindent LEFT -, BACK user backward, TOP user upward and downward: belt, belt buckle, gag, pacifier

\noindent LEFT -, BACK -, TOP user upward: sunhat, beanie, sombrero, cincture, wig, hairnet, dress hat, skullcap, bonnet, shower cap, walking stick, hat, veil, tambourine, life buoy, walking cane, jewel, jewelry

\noindent LEFT -, BACK toward mount backward, TOP -: cufflink, bow decorative ribbons, brake light, motor, broach, blinker, steering wheel, headlight, taillight, reflector, wall socket, latch, doorknob, cleat for securing rope, hinge, propeller, knob, button

\noindent LEFT -, BACK face backward, TOP stand upward: snowman, scarecrow, teddy bear, puppet, doll, rag doll, weathervane, sculpture, mascot, statue sculpture

\noindent LEFT -, BACK -, TOP toward mount upward: tinsel, bell, tassel, pendulum, chandelier, lightbulb, fire alarm, cowbell, punching bag

\noindent LEFT -, BACK away from user, TOP along grab two fingers toward function: nailfile, thermometer

\noindent LEFT -, BACK toward and away from user, TOP along grab one hand: soap, ice pack, butter

\noindent LEFT along grab two hands left right, BACK -, TOP contain or support upward and downward: chopping board

\noindent LEFT -, BACK toward and away from object, TOP stand upward: grater

\noindent LEFT -, BACK user backward, TOP contain or support upward: pipe bowl, tobacco pipe, diving board

\noindent LEFT -, BACK -, TOP object upward: corkscrew, cork bottle plug, bottle cap

\noindent LEFT along grab two hands left right, BACK -, TOP -: rolling pin, barbell, cymbal

\noindent LEFT -, BACK -, TOP cover upward: umbrella, parasol, lampshade, seashell

\noindent LEFT -, BACK away from object, TOP along grab two hands away from function: golf club

\noindent LEFT -, BACK flow backward, TOP along grab one hand toward function: rifle, gun, pistol, water gun

\noindent LEFT -, BACK away from user, TOP stem upward: banana

\noindent LEFT -, BACK toward and away from user, TOP -: bedspread, quilt, towel, blanket, pillow, poker chip, cushion, tarp, dishrag, napkin, cover, bath towel, hand towel, handkerchief, dishtowel, coin

\noindent LEFT -, BACK -, TOP contain or support upward and downward: mattress, hourglass, burrito, taco, coaster

\noindent LEFT -, BACK away from user fingers, TOP user upward: scale measuring instrument, typewriter, computer keyboard

\noindent LEFT -, BACK toward and away from user, TOP toward mount upward: coat hanger, padlock, chime, triangle musical instrument, tea bag, combination lock, curtain, shower curtain

\noindent LEFT -, BACK user forward and backward, TOP user upward: scarf, headset, heart, underwear, legging clothing, suspenders, underdrawers, shawl, diaper

\noindent LEFT -, BACK object squeeze back, TOP along grab two fingers toward function: barrette, safety pin

\noindent LEFT -, BACK toward and away from user, TOP along grab two fingers toward function: plume, popsicle

\noindent LEFT -, BACK -, TOP user upward and downward: headband, sweatband, dog collar

\noindent LEFT -, BACK toward mount backward, TOP toward mount upward: cornice, shower head

\noindent LEFT -, BACK -, TOP move upward: frisbee, balloon, starfish, buoy

\noindent LEFT -, BACK toward mount backward, TOP user upward: poster, wall clock, toilet, handle, knocker on a door, brass plaque, washbasin, kitchen sink, sink, urinal, thermostat

\noindent LEFT -, BACK user lying axis head feet, TOP stand upward: bunk bed, bed

\noindent LEFT -, BACK flow backward, TOP -: spotlight, projectile weapon, telephoto lens, flash, camera lens

\noindent LEFT -, BACK user lying axis head feet, TOP contain or support upward: hammock, sleeping bag

\noindent LEFT -, BACK away from object, TOP toward mount upward: bait, hook

\noindent LEFT -, BACK toward and away from object, TOP along grab one hand: oar, sponge, crowbar, wrench

\noindent LEFT -, BACK toward and away from object, TOP along grab two hands away from function: broom, mop

\noindent LEFT -, BACK away from object, TOP along grab one hand: scrubbing brush

\noindent LEFT -, BACK toward and away from object, TOP toward mount upward: clasp

\noindent LEFT -, BACK away from object, TOP user upward: drill

\noindent LEFT -, BACK toward mount backward, TOP user upward and downward: pegboard

\noindent LEFT -, BACK away from object, TOP along grab two fingers toward function: key, candy cane

\noindent LEFT -, BACK -, TOP along grab two hands away from function: spear

\noindent LEFT -, BACK away from user fingers, TOP stand upward: violin, guitar

\noindent LEFT -, BACK entry backward, TOP contain or support upward: birdhouse, birdcage

\noindent LEFT -, BACK entry backward, TOP stand upward: kennel

\noindent LEFT -, BACK user forward and backward, TOP contain or support upward: stirrup

\noindent LEFT -, BACK -, TOP toward and away from mount backward: deadbolt

\noindent LEFT -, BACK away from user, TOP -: earplug, band aid, compass

\noindent LEFT along grab two hands left right, BACK -, TOP eyelet axis: handcuff

\noindent LEFT -, BACK user backward, TOP -: shield

\noindent LEFT along grab two hands left right, BACK -, TOP stand upward: griddle

\noindent LEFT -, BACK flow backward and forward, TOP toward mount upward: wind chime

\noindent LEFT -, BACK -, TOP stand upward downward: crawfish, boiled egg, egg, fish food, salmon fish, pancake

\noindent LEFT -, BACK -, TOP stem upward and downward: lime, watermelon, lemon, kiwi fruit, melon

\noindent LEFT -, BACK hand backward and forward, TOP hand fingers: potholder, mitten, glove

\noindent LEFT -, BACK move backward, TOP move upward and downward: kite

\noindent LEFT -, BACK toward and away from user, TOP contain or support upward and downward: matchbox

\noindent LEFT -, BACK toward and away from mount backward, TOP -: wheel

\noindent LEFT -, BACK away from user, TOP user upward and downward: receipt

\noindent LEFT -, BACK -, TOP cover upward and downward: runner carpet, doormat

\section{Manual Labor}
\label{sec:sup_manual_labor}

Despite manually annotating only 1K 9D object poses, we verify all oriented objects in the dataset. On average, each 9D pose annotation requires approximately 4 minutes, while verification of a pose takes around 2 seconds. This corresponds to about 67 hours of human effort for manual 9D pose annotation and 132 hours for large-scale pose verification. In total, the dataset curation required approximately 200 hours of human labor.

\begin{figure}
    \centering
    \includegraphics[width=\linewidth]{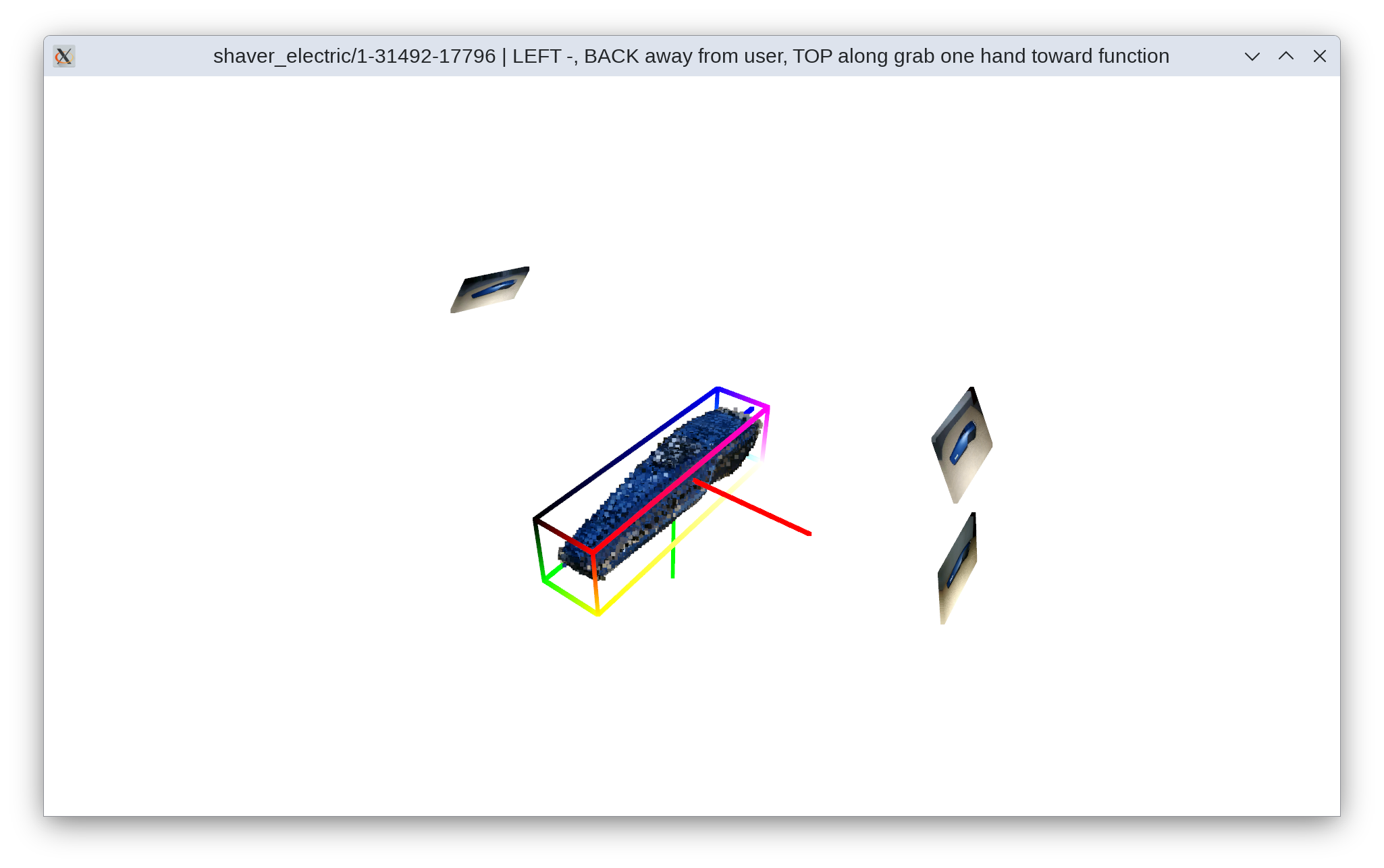}
    \caption{\textbf{Annotation Tool.} To annotate reference objects, we developed an interactive tool that enables precise $1^\circ$ adjustments of the current axes while visualizing the oriented bounding box and canonical axes in real time. The tool also supports switching between different object views, which facilitates accurate refinement of the canonical frame.} 
    \label{fig:sup_method_annotation9d}
\end{figure}

\begin{figure}
    \centering
    \includegraphics[width=\linewidth]{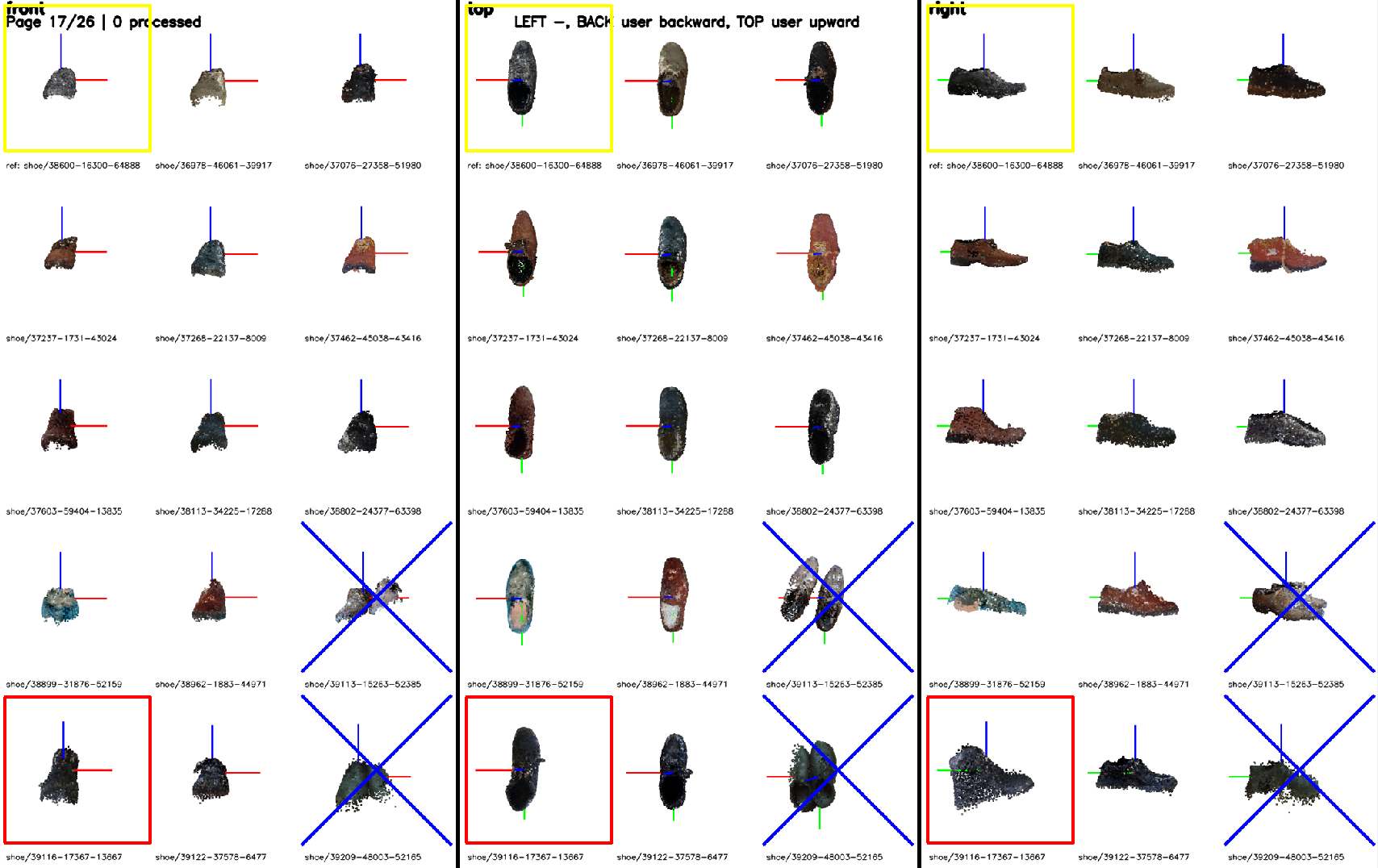}
    \caption{\textbf{Verification Tool.} To verify objects, we render them from three views, front, top, and right. This enables us to discover fast and precise not aligned object instances and filter out objects due to bad reconstruction or multiple objects.}
    \label{fig:sup_method_verification9d}
\end{figure}

We illustrate the annotation interface in \cref{fig:sup_method_annotation9d} and the verification interface in \cref{fig:sup_method_verification9d}.

\clearpage

\section{Qualitative Examples}
\label{sec:sup_qualitative_examples}

\begin{figure} %
\centering
\caption{\textbf{Overview of Every9D-22M.}}
\includegraphics[width=0.78\linewidth]{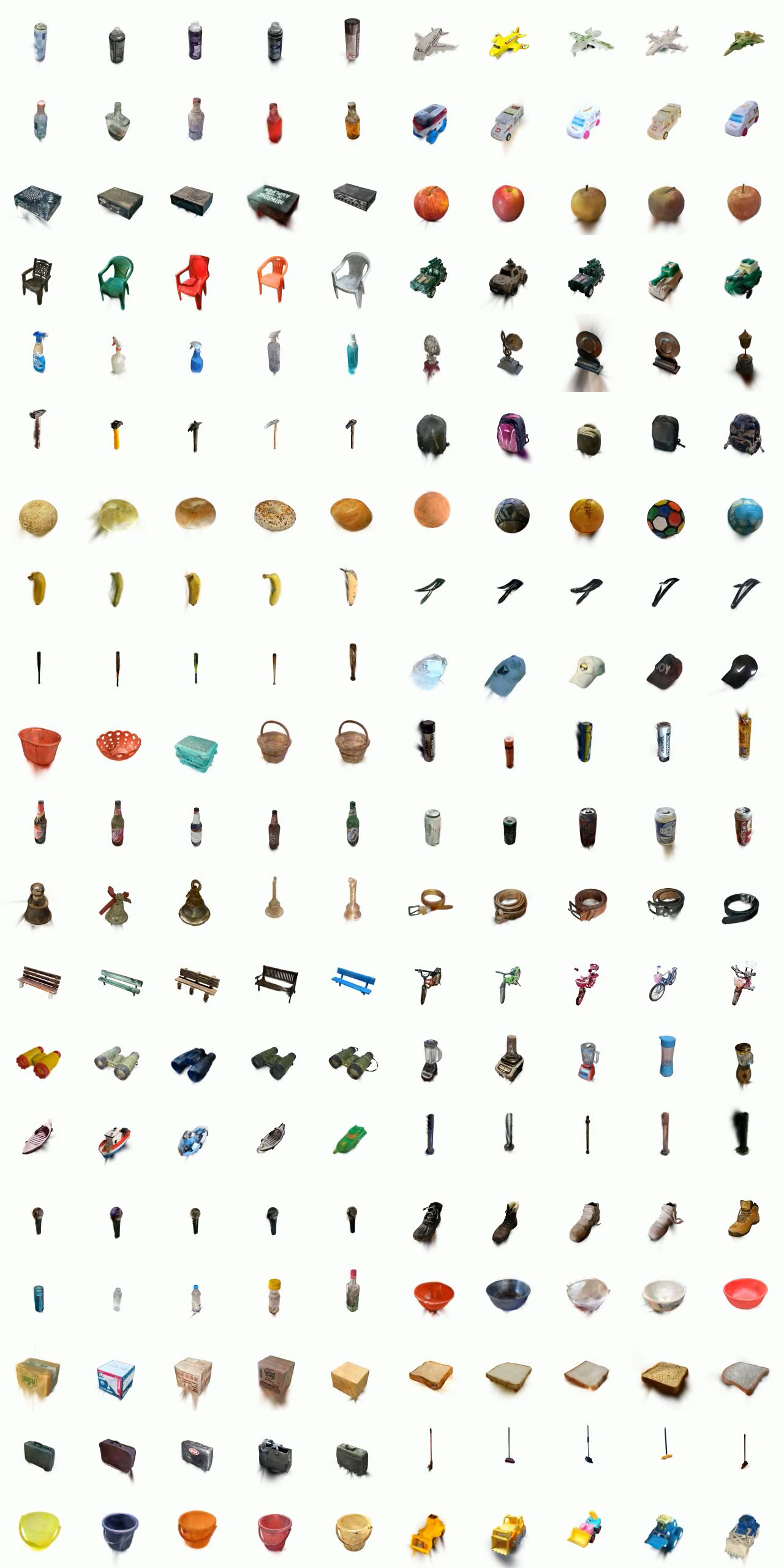}
\label{fig:sup_qualitative_examples_vis}
\end{figure}

We provide a qualitative overview of 3D Gaussian Splatting reconstructions for a sample of 1,000 canonicalized objects spanning 200 categories. The examples illustrate the diversity of the dataset and the consistency of the resulting canonical object frames across instances and categories, see \cref{fig:sup_qualitative_examples_vis}.

\begin{figure} %
    \centering
    \includegraphics[width=0.78\linewidth]{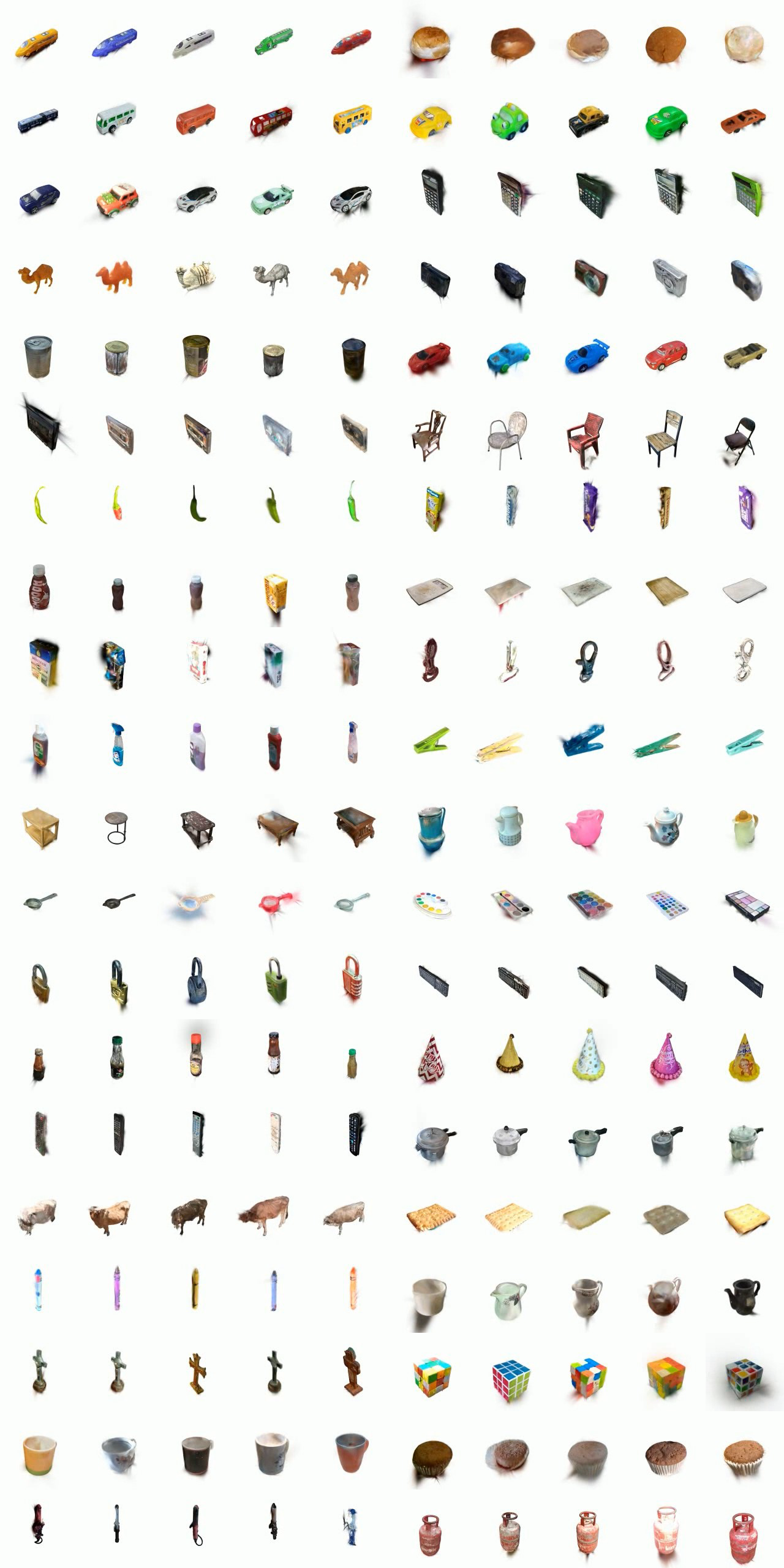}
\end{figure}

\begin{figure} %
    \centering
    \includegraphics[width=0.78\linewidth]{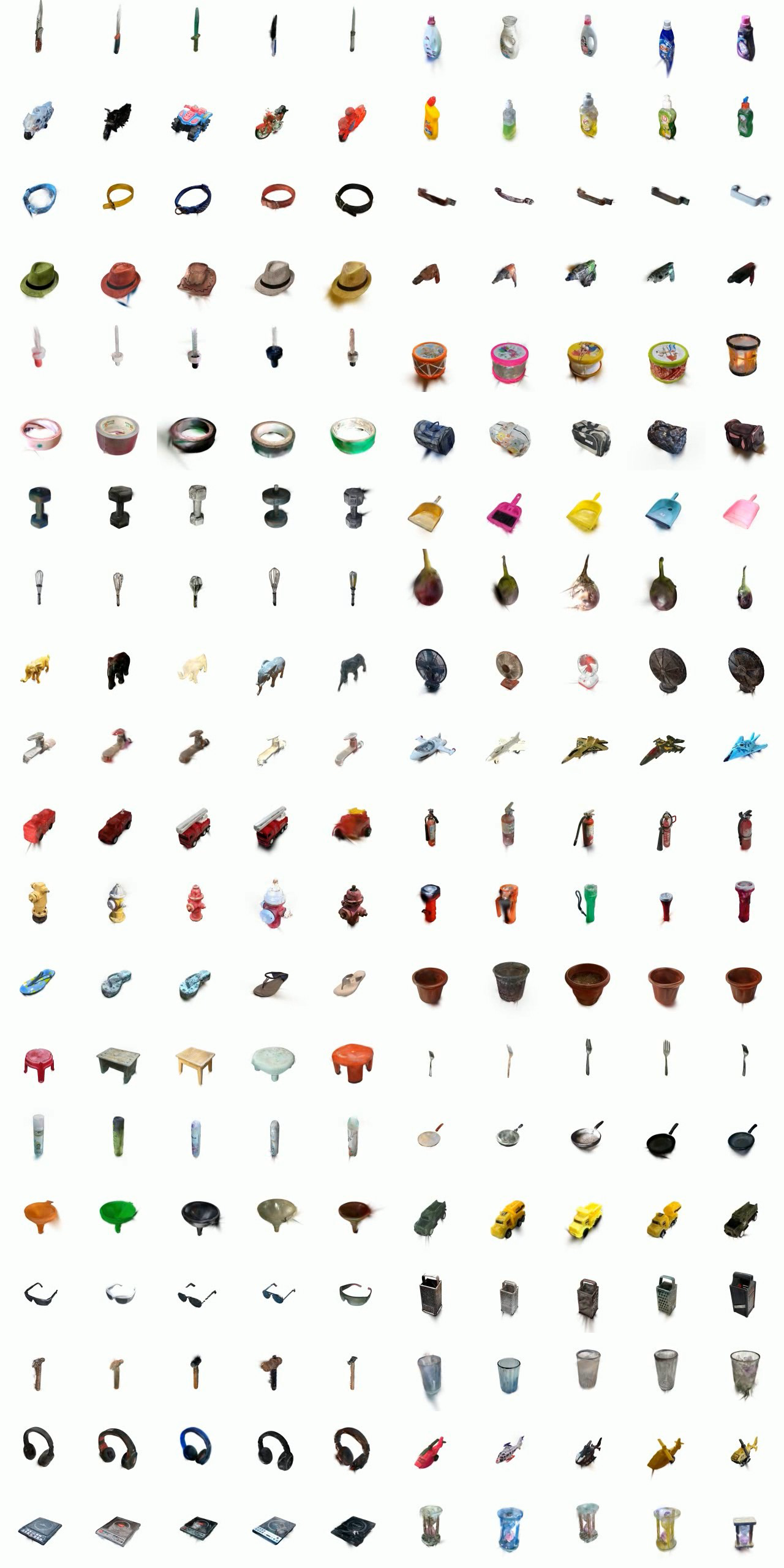}
\end{figure}

\begin{figure} %
    \centering
    \includegraphics[width=0.78\linewidth]{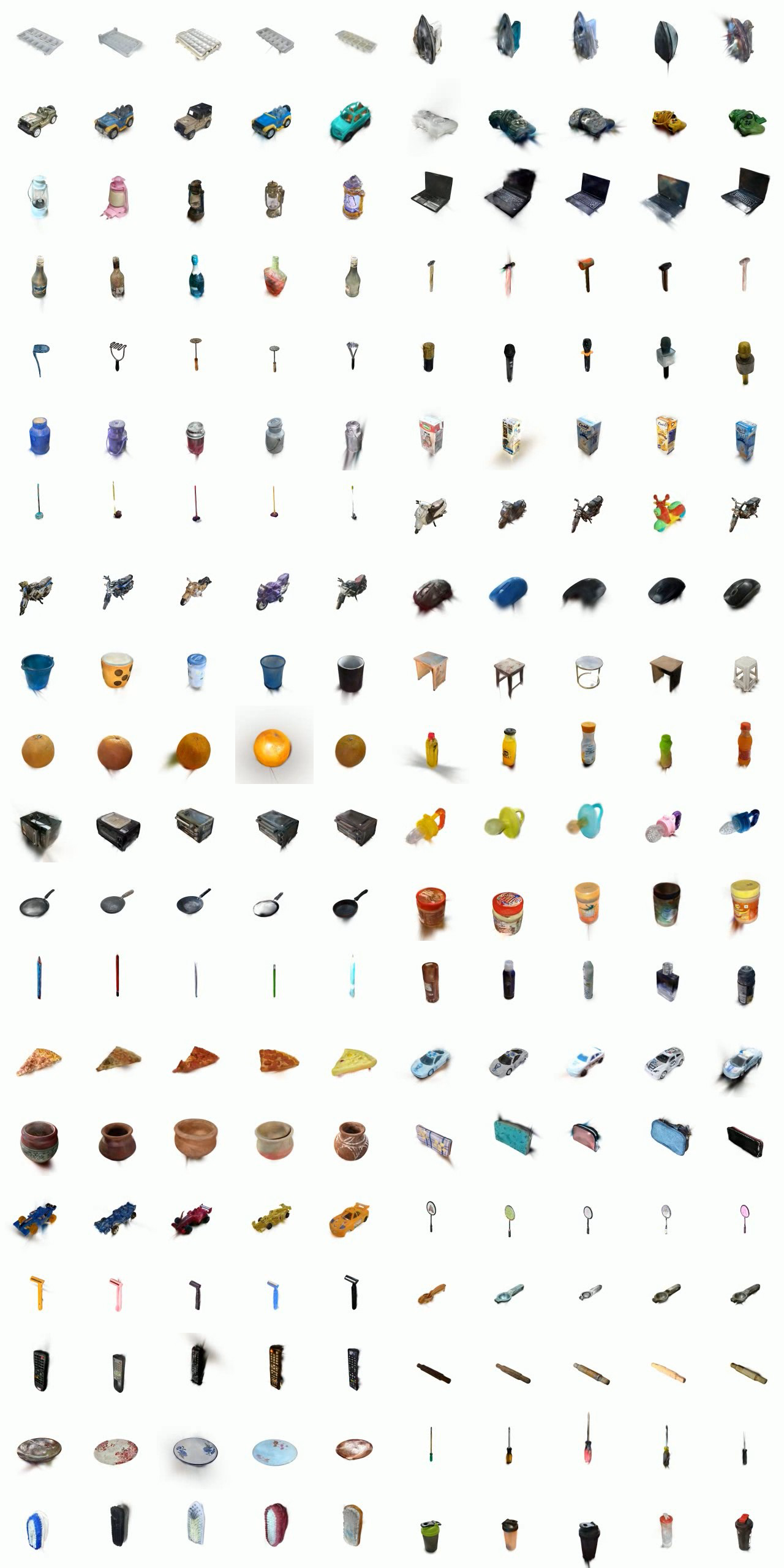}
\end{figure}

\begin{figure} %
    \centering
    \includegraphics[width=0.78\linewidth]{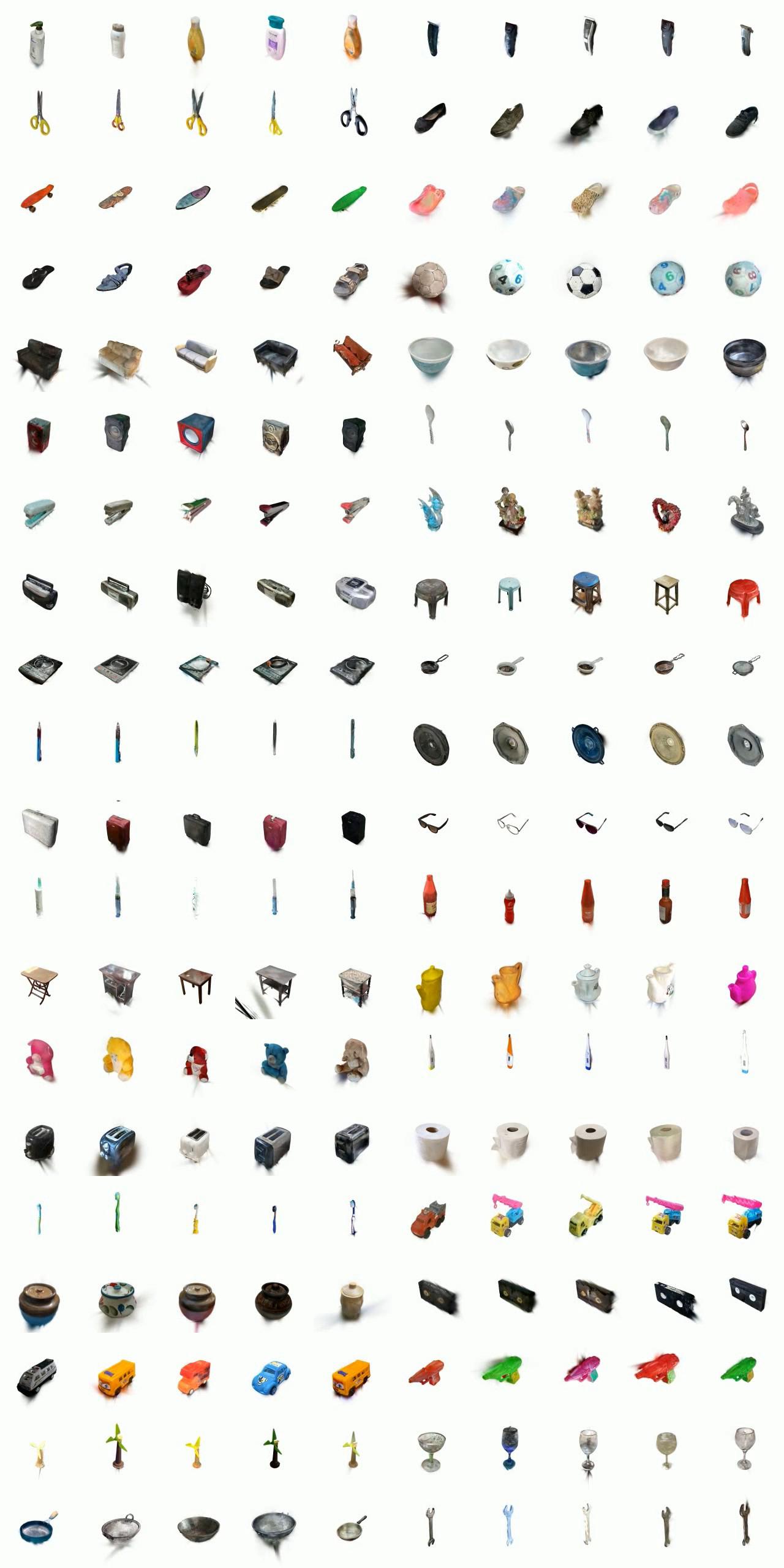}
\end{figure}

\section{Societal Impact}
\label{sec:sup_societal_impact}
Large-scale 9D pose estimation has the potential to advance applications in robotics, AR/VR, embodied AI, and 3D content creation by enabling systems to reason about object geometry, orientation, and physical interaction from visual observations. Improved object-centric spatial understanding may benefit assistive robotics, warehouse automation, autonomous manipulation, digital twin construction, and immersive virtual environments. Furthermore, scalable canonicalized real-world datasets can reduce reliance on synthetic data and support more robust and generalizable geometric foundation models.

At the same time, large-scale 9D pose estimation systems may introduce risks when deployed in real-world settings. Improved object tracking and scene understanding could contribute to surveillance applications or enable more capable autonomous systems in sensitive domains. Dataset biases, including imbalances in object categories, geographic distribution, or environmental conditions, may also lead to uneven performance across deployment settings. In addition, reliance on monocular depth estimation rather than sensor depth can propagate geometric inaccuracies, potentially affecting downstream safety-critical applications.

We mitigate some of these concerns by focusing on object-centric videos of everyday objects rather than human-centered data, and by releasing standardized evaluation protocols to encourage transparent benchmarking. Nevertheless, we emphasize that 9D pose foundation models should be deployed with appropriate safeguards, particularly in safety-critical or privacy-sensitive environments. Future work should further investigate fairness, robustness, uncertainty estimation, and mechanisms for preventing misuse in surveillance or autonomous decision-making systems.

\clearpage

\end{document}